\documentclass[lettersize,journal]{IEEEtran}
\usepackage[numbers]{natbib}
\usepackage[utf8]{inputenc}
\usepackage{tikz}
\usepackage{pgfplots}
\usepackage{tikz-3dplot}
\usepackage{textcomp}
\usepackage{lineno}
\usepackage{booktabs}
\usepackage{balance} 
\usepackage{tabularx}
\usepackage{mathtools}
\usepackage{graphicx}
\usepackage{graphics}
\usepackage{epstopdf}
\usepackage{amsmath,amssymb,amsfonts}
\usepackage[english]{babel}
\usepackage{amsthm}
\usepackage{soul}
\usepackage{color}
\usepackage{subcaption}
\usepackage{caption}
\usepackage{wrapfig}
\usepackage{comment}
\usepackage{float}
\usepackage{amssymb}
\usepackage{xcolor}
\usepackage{lipsum}
\usepackage{enumitem}
\usepackage{bbm}
\usepackage{hyperref}
\usepackage{pdflscape}
\usepackage{natbib} 
\usepackage{algorithm}
\usepackage{algpseudocode}

\begin{document}

\title{Artificial Neural Networks-based\\ Real-time Classification of ENG Signals\\ for Implanted Nerve Interfaces
}

\author{Antonio Coviello~\IEEEmembership{Student Member,~IEEE}, Francesco Linsalata~\IEEEmembership{Member,~IEEE},\\ Umberto Spagnolini,~\IEEEmembership{Senior Member,~IEEE}, Maurizio Magarini,~\IEEEmembership{Member,~IEEE}, \\ 
mail to: antonio.coviello@polimi.it
\thanks{The authors are with the Dipartimento di Elettronica, Informazione e Bioingegneria, Politecnico di Milano, 20133 Milan, Italy.}}
\maketitle

\begin{abstract}
Neuropathies are gaining higher relevance in clinical settings, as they risk permanently jeopardizing a person's life. To support the recovery of patients, the use of fully implanted devices is emerging as one of the most promising solutions. However, these devices, even if becoming an integral part of a fully complex neural nanonetwork system, pose numerous challenges. In this article, we address one of them, which consists of the classification of motor/sensory stimuli. The task is performed by exploring four different types of artificial neural networks (ANNs) to extract various sensory stimuli from the electroneurographic (ENG) signal measured in the sciatic nerve of rats. Different sizes of the data sets are considered to analyze the feasibility of the investigated ANNs for real-time classification through a comparison of their performance in terms of accuracy, F1-score, and prediction time. The design of the ANNs takes advantage of the modelling of the ENG signal as a multiple-input multiple-output (MIMO) system to describe the measures taken by state-of-the-art implanted nerve interfaces. These are based on the use of multi-contact cuff electrodes to achieve nanoscale spatial discrimination of the nerve activity. The MIMO ENG signal model is another contribution of this paper.
Our results show that some ANNs are more suitable for real-time applications, being capable of achieving accuracies over $90\%$ for signal windows of $100$ and $200\,$ms with a low enough processing time to be effective for pathology recovery.
\end{abstract}

\begin{IEEEkeywords}
Electroneurographic (ENG) Signal, Neural Engineering, Neural Networks Classifications, Peripheral Nerve Interface.
\end{IEEEkeywords}

\section{Introduction}
\label{sec:intro}

The peripheral nervous system (PNS) can be compromised due to several causes, leading to peripheral neuropathies (PNs). Treatment of PNS lesion is still an unmet clinical and scientific need. PNs can be characterized by sensory, motor, and autonomic disorders in the affected areas. They may involve a single nerve (mononeuropathy), two or more different nerves in different locations (multiple mononeuropathy), or there may be a widespread, often symmetrical, bilateral process (polyneuropathy) \cite{[1]lehmann2020diagnosis, [2]castelli2020peripheral, [3]girach2019quality}. 

Peripheral neuropathies affect $2\,$-$\,3\%$ of the population and seriously interfere with quality of life. People with PN experience reduced quality of life and the socio-economic impact of residual disability is potentially very significant: even simple tasks (e.g., buttoning a shirt) can become unattainable or overly burdensome. Mechanical injuries are frequent with $2.8\%$ of traumatized patients \cite{ [4]caillaud2019peripheral, [5]Denison65, [2a]grill2009implanted}. As a result of a peripheral lesion there is a phenomenon of degeneration of nerve fibers. Although regeneration is possible, it could not take place correctly, thus impairing the functional recovery. For this reason, traditional medicine has sought alternative methods to make up for this lack \cite{[5]Denison65, [2a]grill2009implanted}. 

Promising therapeutic interventions based on the use of bioelectronic medicine depend on specific implantable devices of peripheral nerve interfaces (PNIs)~\cite{[1a]LARSON2020108523, [8]russell2019peripheral}. Advances in MEMS technologies and the development of miniaturized electronic systems now make it possible to realize smaller devices that can be implanted inside the body \cite{[5]Denison65}. PNIs are implantable devices used to monitor and modulate peripheral nerve activity by measuring the electroneurogram. Nowadays, neural decode and stimulate (ND\&S) systems that record multiple nerve signals, decode individual activity, and relay them for stimulation are an open research area \cite{[6]Elisa, [7]Federica, [5]Denison65}. Therefore, PNIs able to sample residual PNS signals might have a primary role in the realization of ND\&S that support the restoration of lost functionalities. 

\begin{figure*}[!t]
\centering
\subfloat[]{\includegraphics[width=0.7\columnwidth]{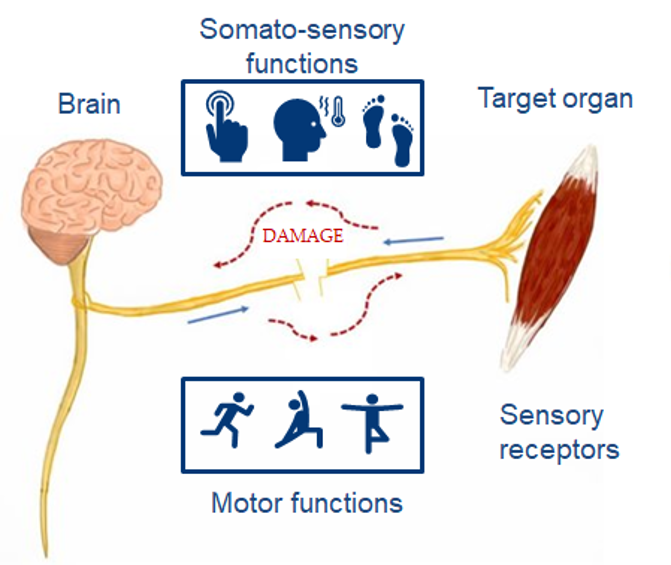}%
\label{fig_first_case}}
\hfil
\subfloat[]{\includegraphics[width=0.7\columnwidth]{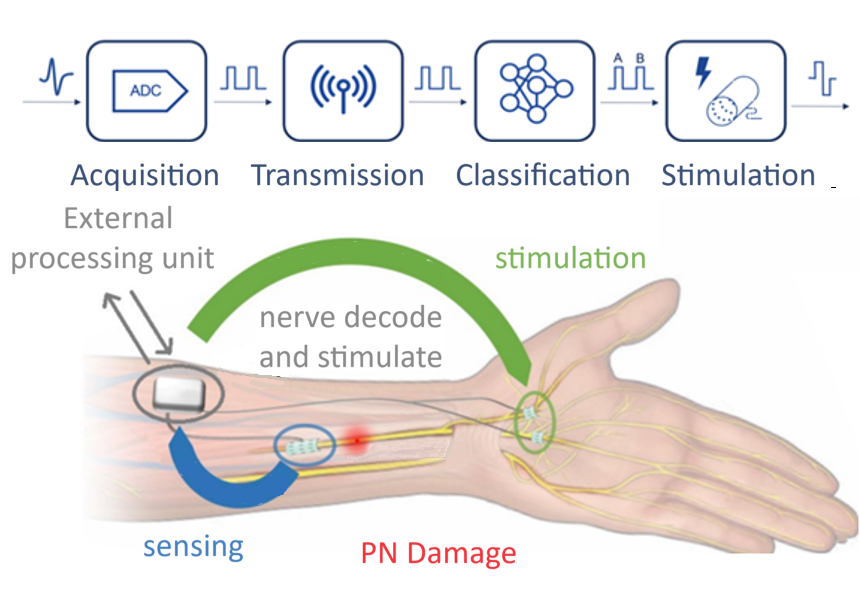}%
\label{fig_second_case}}
\caption{(a) Representation of a nerve injury and related issues. (b) Steps necessary to re-establish the connection \cite{[6]Elisa} and their possible implementation within the body \cite{[7]Federica}.}
\label{fig1}
\end{figure*}

A main challenge for the development of ND\&S systems is the ability to classify in real-time electroneurographic (ENG) signals~\cite{[5]Denison65}. This is mainly due to the lack of classification approaches that can account for the aggregate propagation mechanism in nerve's axons. In addition to this, when carrying out the analysis of biological signals, the inherent delay introduced by the processing must be also considered. When it comes to implanted devices for the restoration of sensory/motor functions, any delay caused by the data processing becomes an important element that must be taken into account. In fact, it must be considered that the human response time is about $300\,$ms \cite{[10][11]Controller}, and consequently for real-time operations it is necessary that all the processing and retrieval must be carried out within this time interval constraint.
 
The nervous system is an electromagnetic nanonetwork  facilitating communication at nanoscale levels. While electromagnetic nanonetworks rely on electromagnetic waves for information exchange within wireless communication systems, the nervous system employs neurons as nanotransceivers to transmit signals between the brain and the periphery, or vice versa. These neurons form a large-scale nanonetwork spanning the entire body, wherein electrical signals excite them to carry out functions such as storing, processing, and transmitting information through both chemical and electrical signaling mechanisms \cite{EMnanocomunications, malak2014communication}.
Understanding and describing compactly, as a nano networked system, the propagation of this information within the body is fundamental, especially if we consider the numerous axons contained in the nerves.

\textbf{Contributions: }
In this particular context, it is clear that the development of a well-defined system architecture capable of real-time stimulating the intended sensory/motor stimulus 
would represent a great innovation. 
Accordingly, the primary contribution of our work is summarized by the following key points:
\begin{itemize}
\item Definition of an ND\&S architecture that measures ENG signals using a PNI, decodes them by appropriate pre-processesing followed by ANN-based classification, and forwards the extracted intended sensory/motor stimulus after the nerve injury;
\item Definition of a multiple-input multiple-output (MIMO) ENG model that accounts for the aggregate propagation of the nerve motor and sensory signal nano activities. The model derives from the deployment of highly spatial-selective multi-contact cuff electrodes to measure the ENG signals in state-of-the-art PNIs. A multi-contact cuff is currently not able to probe selectively different axons and the MIMO ENG is designed for this purpose;
\item Design of artificial neural network (ANN) approaches suitable for real-time classification of ENG signals. The design takes into account the MIMO ENG signal propagation model within temporal windows of $300\,$ms, compatible with the human response time;
\item Comparison of the performance for the developed ANN-based classification methods through their application on real measured ENG data sets. These are available public data sets containing four types of sensory signals measured in the sciatic nerve of three rats \cite{[12]Identification}
\item Evaluation of performance metrics, i.e. accuracy, F1-score, and test time. This comprehensive analysis provides insights into the effectiveness and efficiency of our architecture and modelling in different conditions.
\end{itemize}
Through the above contributions, our work not only addresses the current void in the system architecture and modelling of ND\&S, but also highlights the boundaries of real-time classification of ENG signals, fostering a deeper understanding of their practical implications.

\textbf{Organization:} The rest of the paper is organized as follow. Section II describes the PNIs and their deployment in the proposed ND\&S architecture. Section III summarizes the state of the art of peripheral nerve signal classification. The proposed MIMO ENG signal model suitable for the development of ANN-based classifiers is defined in Sec. IV. Section V exploits this model for different ANN architectures. Their performances over real measured MIMO ENG signals from the sciatic nerves of rats are reported in Sec. VI, while the differences in performance with state of the art solution is highlighted in Sec. VII. Section VIII summarizes and concludes the work.

\section{Peripheral Nerve Interfaces in ND\&S Systems}
\label{sec:NDS}

Figure~\ref{fig1}a gives a pictorial description of a peripheral nerve injury and the main related consequences. Injury translates into the interruption of:
\begin{itemize}
    \item \textit{efferent signal}, from the brain to the body, preventing any effective action of the brain;
    \item \textit{afferent signal}, from the periphery of the body to the brain, an important signal to ensure a controlled feedforward system within the body.
\end{itemize}

Nerve fibers may regenerate spontaneously, but it is slow (approx $0.4-2\,$mm/day) and depends on various factors such as age and severity~\cite{[5a]SIEMIONOW2009141}. Surgery for nerve regeneration has shown that only about $40$\% achieve satisfactory sensory and motor recovery, respectively~\cite{[7a]grinsell2014peripheral}. Both afferent and efferent pathways are necessary to restore a patient-controlled motor act, particularly of the feedforward type. The main idea that motivates the development of an ND\&S system is that when axonal degeneration occurs, regeneration almost always follows, although often insufficient to guarantee complete functional recovery. However, if there is a minimal population of regenerated axons in which ENG is partially transmitted, this residual activity can be captured \cite{[4]caillaud2019peripheral, [5]Denison65, [2a]grill2009implanted}. 

An ND\&S system can be implemented as a whole when the nerve signals of the axons can be measured, separated, and classified. This is possible only when the ENG signal is measured by means of a multi-contact cuff electrode~\cite{[1a]LARSON2020108523,[8]russell2019peripheral}. To ensure a feedforward system, the ND\&S system must restore both motor and sensory functions. The result of the classification is used to generate the signal required for stimulation. In case of motor functionality, the decision command extracted from the classifier is sent to actuators located in proximity of the muscle and through a controlled stimulation directed at the level of the muscles~\cite{[5]Denison65}. 
\begin{figure}[!t]
\centering
\includegraphics[width=1\columnwidth]{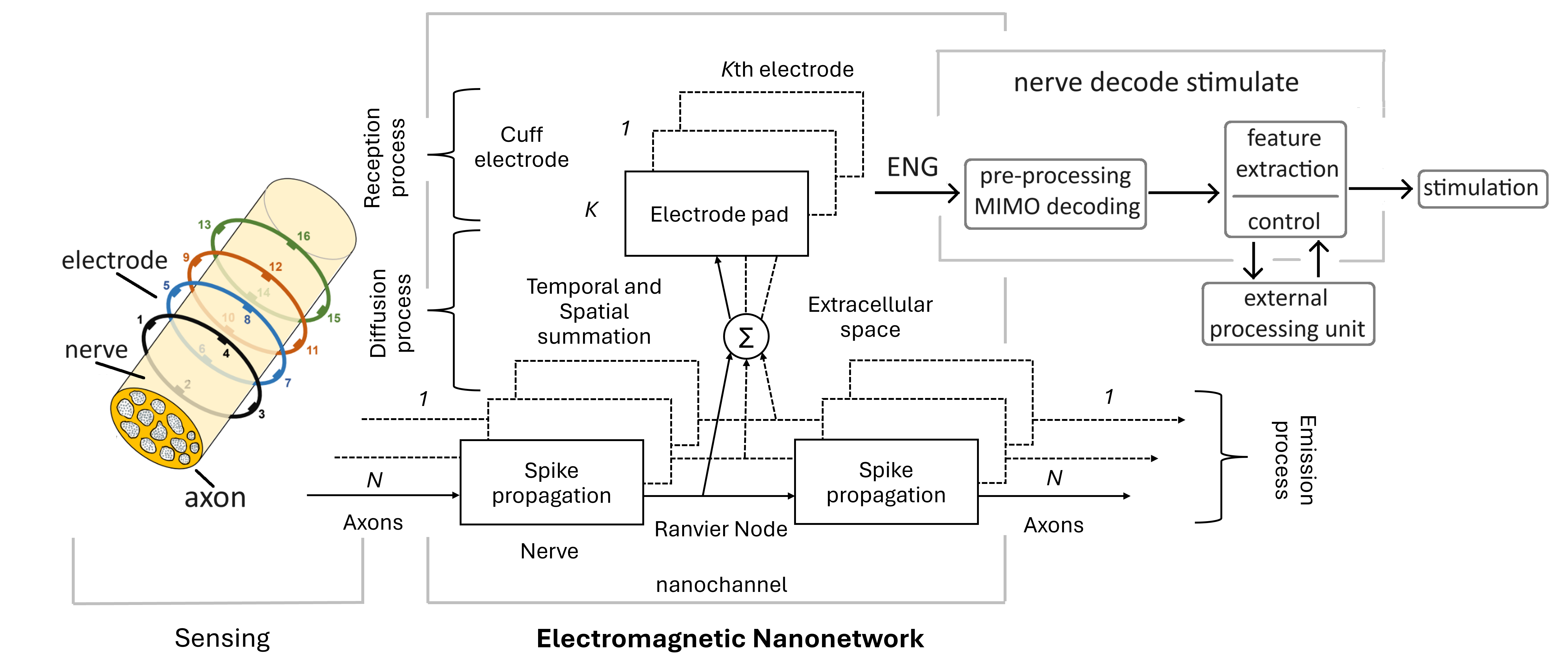}
\caption{The electromagnetic nanonetwork of the nerve/axon-cuff electrode interface. From the ENG measurement to the data processing.}
\label{fig2a}
\end{figure}
\begin{figure}[!t]
\centering
\includegraphics[width=0.9\columnwidth]{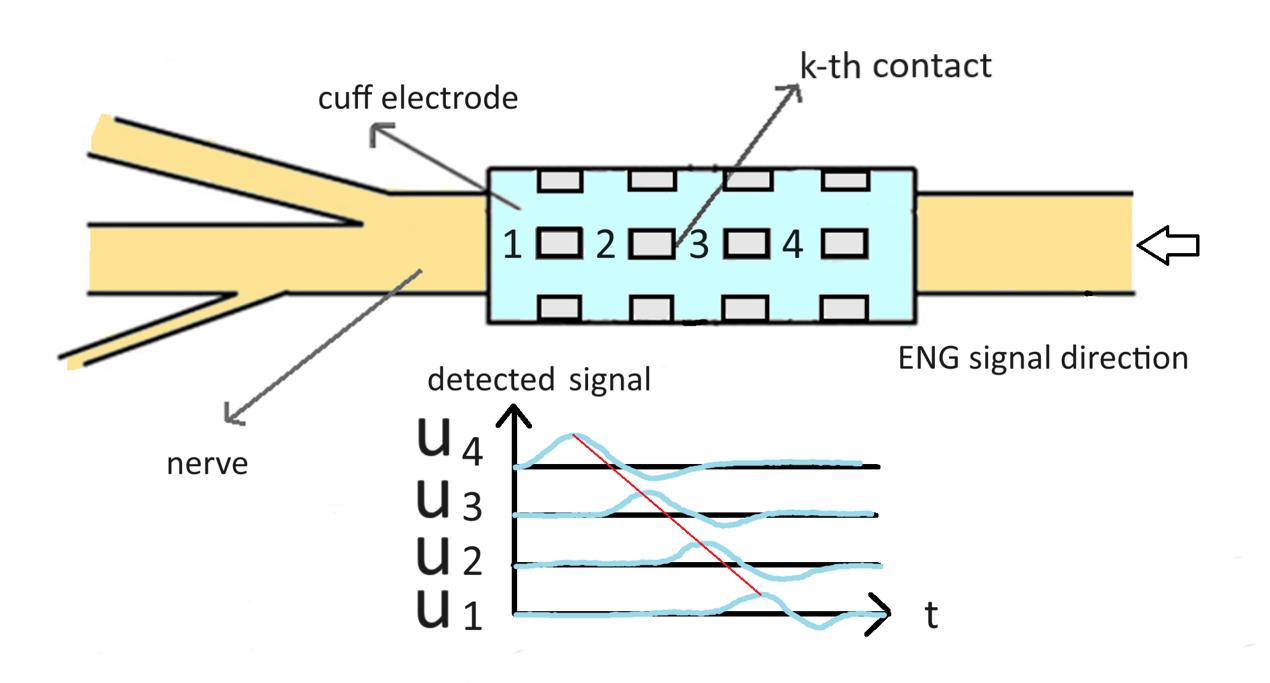}
\caption{Sampled signal along the contacts of the cuff electrode.}
\label{fig1b}
\end{figure}

Figure~\ref{fig1}b gives a pictorial example of a complete bidirectional ND\&S system to restore motor and somato-sensory functions~\cite{[6]Elisa, [7]Federica}. The reported scheme details the main steps involved in the decoding and forwarding nerve signals for the reestablishment of the physiological pathways. The process starts with the acquisition of the ENG signal by means of a multi-contact cuff electrode. This is followed by an analog-to-digital conversion (ADC)~\cite{[8]russell2019peripheral,[1a]LARSON2020108523} and by a pre-processing to reduce the effect of the noise and of the distortions. The pre-processed signal is finally transmitted to an external processing unit (PU) for the classification task. 

The external transfer to the PU is necessary due to the limited computational capabilities of the actual implanted devices, which typically comprise micro-controllers with restricted storage and power supply constraints~\cite{[5]Denison65}. The transmission outside the body (uplink) utilizes Bluetooth Low Energy (BLE) as for the communication protocol. After classification, the signal is sent back to the implanted device for neuromodulation (downlink), ultimately supporting the restore of the lost functionality~\cite{[5]Denison65,[6]Elisa, [7]Federica}. The multi-contact cuff electrode is positioned around the nerve, as shown in Fig.~\ref{fig2a}. This is arranged in a matrix-like configuration and extracts both nerve spatial (transversal) and temporal (longitudinal) information, as depicted in Figure~\ref{fig1b}.

\begin{figure}[!t]
\centering
\includegraphics[width=0.9\columnwidth]{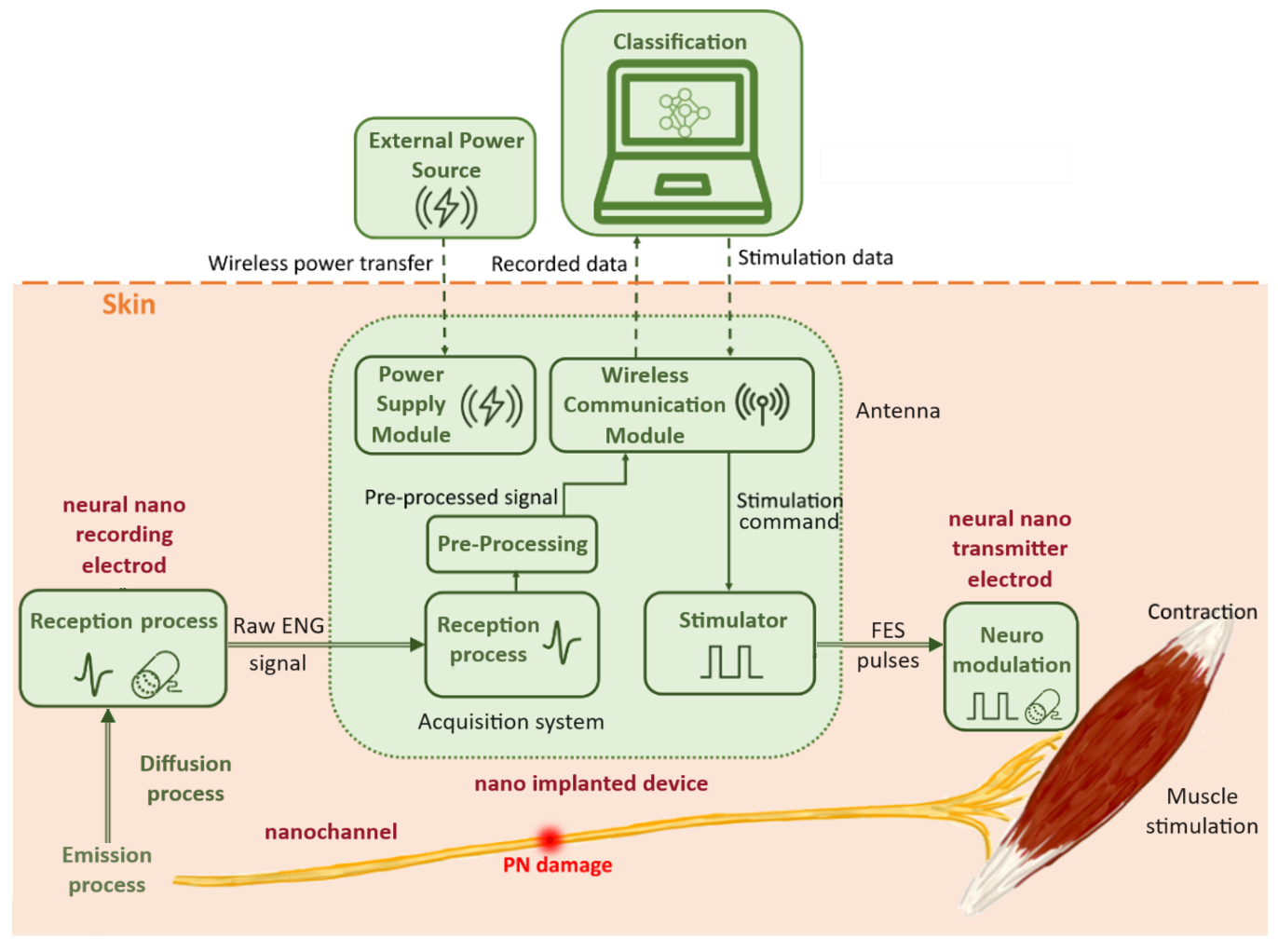}
\caption{Detail of the all operations implemented in a ND\&S system to implement a functional neural bypass.}
\label{fig4a}
\end{figure}

Regarding the latency introduced in the communication pipeline, a crucial parameter an important parameter to be considered in ND\&S systems is the size of the data payload, which is proportional to the temporal duration $T_w$ of the window used for the classification. Besides $T_w$, the size of the payload also depends on the number of contacts of the cuff electrode, the chosen sampling frequency, and the ADC bit resolution. Considering all these aspects, the size of the payload is given by
\begin{align} \label{eq:Pay}
    P= f_s \, N_{ch}\, \,T_w \,b, \, \,\,\,\ [\text{bit}]
\end{align}
where $f_s$ is the sampling frequency, $N_{ch}$ is the total number of electrodes around the nerve, and $b$ is the ADC resolution (in bits). The size of the payload impacts another important parameter, which is the overall latency introduced by the sequence of operations implemented by the ND\&S system. To evaluate the overall latency we can refer to Fig.~\ref{fig4a}, which shows a complete ND\&S system where each block introduces a delay. The total feed-forward time $T_f$ is given by the sum of different contributions as follows:
\begin{align} \label{eq:Tf}
    T_f= T_a+T_w+T_u+T_c+T_d+T_n, \,\,\,\,[\text{ms}]
\end{align}
where $T_a$ is the acquisition time, $T_u$ is the uplink transmission time, $T_c$ is the classification time, $T_d$ is the downlink transmission time, and $T_s$ is the neuro-modulation time.
$T_f$ must be lower than $300\,$ms for real-time applications. This limit sets a constraint on the delay introduced by the classifier \cite{[10][11]Controller}. 

\section{Related Work}
\label{sec:sda}
Achieving almost real-time classification is of fundamental importance for the advancement and future deployment of ND\&S systems. Given that the human physiological response needs a timely classification of the neural signal to avoid a deficit perception, the extraction of appropriate features from the measured nerve signal becomes a crucial aspect for the classification. In the literature, three main types of approaches have emerged for the analysis of ENG signals, which are illustrated hereafter:

\subsection{Spike detection} 
The aim of this category of approaches is to extract the cardinal building block of an ENG signal, i.e., the \textbf{action potential (AP)} (also termed as ``\textbf{spike}'')~\cite{[14]koh2020selective}. In this method the classification is applied only to the part of measure where a spike is detected. In order to obtain a good spike detection performance, the signal must be very clean. This happens when the signal-to-noise ratio is high ($\geq 10\,$dB). Therefore, to implement the algorithm an effective de-noising of the signal is necessary. In \cite{[14]koh2020selective} this is done by pre-filtering of the noisy measured signal followed by a thresholding for spike detection. A convolutional neural networks (CNN) is then used to classify the spatio-temporal signatures built from the AP detected inside the ENG signal. A group of $3$ different classes from different mechanical sensorial stimuli is considered. It is shown that $80\,$\% in mean accuracy and $75\,$\% in F1-Score are achieved for the considered dataset at hand.

\subsection{Feature extraction using moving observation windows} In this class of methods, the extraction of classification features from the measured signal is done through the definition of a temporal window with length that, in principle, should be well below the $300\,$ms limit defined by the human response. This allows to obtain a general characterization of the measured signals in terms of nerve activity in the window. Such a class of techniques is suitable for measures characterized by low signal-to-noise ratio ($\leq 10\,$dB). In~\cite{newcastle2020}, a feature extraction framework is performed using moving observation windows and different types of machine learning techniques. The used dataset concerns with six different levels of intensity using the same mechanical sensorial stimuli from the sciatic nerve of rats. Different window sizes between $50$ and $350\,$ms have been used. Excellent classification results are shown with accuracy values above $90$\%.
\subsection{Unsupervised hybrid kernel classification} 
This category is a hybrid of the previous two. Actually, it combines the measure of the activity in a pre-defined temporal window with a specific sensorial stimulus. The latter is defined in terms of a local information, given by a spike with a certain shape, and a global one, given by the spike temporal density. This observation allows us to introduce kernels to improve the extraction of information. In fact, kernels are able both to implement the recognition of APs, i.e. local information, and to extract the global information, i.e. the frequency of appearance~\cite{[6]Elisa,[7]Federica}. Note that, both spikes and global features have to be considered in the classification. Therefore, the hybrid method seems to be the most suitable to have a balance between local and global information. In \cite{[7]Federica}, a CNN with two 1D filters is employed to extract information from ENG signals. The used dataset contains $10$ different mechanical sensorial stimuli. It is shown that the CNN achieves a mean accuracy of $90.6\,$\%, and an F1-Score of $89.9\,$\% using a window of $1\,$s extracted from the original signal. Meanwhile, a novel ConvLSTM has been developed in \cite{[6]Elisa} to classify the same dataset used in \cite{[7]Federica}. In this case the dataset where grouped into $4$ main classes. Samples of the original signal are used with windows of duration from $100\,$ms up to $2.5\,$s. Results show a mean in F1-Score of $91.15$\%$\pm 4.36$\%.

Due to its superior performance compared to the first two methods, in this paper we have decided to use the hybrid approach.
\begin{figure}[!t]
\centering
\includegraphics[width=0.8\columnwidth]{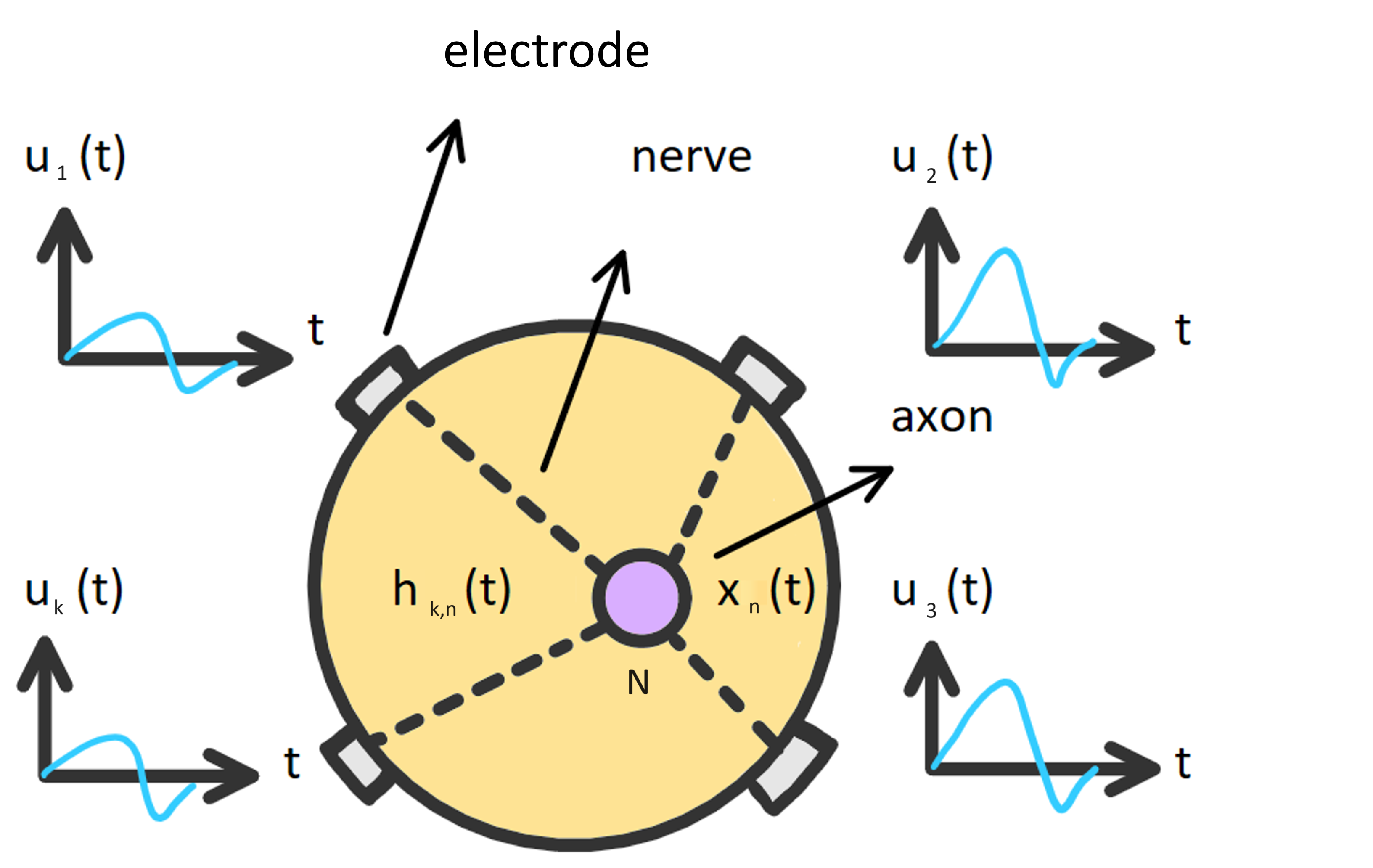}
\caption{Section of a peripheral nerve showing the $n$th axon disposed inside the nerve generating a signal $x_n(t)$. }
\label{fig2b}
\end{figure}

\section{Proposed MIMO Electroneugraphic (ENG) Signal Model}
\label{sec:model}
The model of the peripheral nerve used to design the ANN-based classification approaches described in this paper is delineated in Fig.~\ref{fig2b}. The figure shows a section of the nerve where the $n$th axon is highlighted, with $n=1,\ldots,N$, being $N$ the number of considered axons\footnote{Actually, the number of individual axons is unknown. However, without loosing of generality, in the present analysis it is assumed to be known since it can be estimated as detailed in~\cite{mishra2022modeling}.}.
Each axon controls a specific set of muscles for movement. The cuff electrode measures the aggregated nerve signals propagating in each of the individual axons~\cite{[1a]LARSON2020108523, [8]russell2019peripheral}. The possibility of recovering each of the nerve signals from the measured mixture is of utmost importance for precise movement recovery. Hereafter, as one of the main contributions of this work, we introduce a model that takes into account the presence of multiple contacts on the cuff electrode and of the mechanisms at the basis of the propagation of the electroneugraphic signals. 

To describe the ENG signal $x_n(t)$ we need to start from its constituent element, i.e., the spike. This is the result of a rapid change of the transmembrane potential in the $n$th axon that allows for the propagation of information within the nervous system~\cite{assone}. When observing a single fiber, the associated waveform $x_n(t)$ can change depending on the activated neuron. 
A spike is typically modelled as
a unipolar voltage generated by the activity of the $n$th axon ~\cite{[1b]TaylorSummary}:
\begin{equation}
    x_n(t)=    \left\{\begin{split}
A_nt^{m_n}e^{-B_nt} &  \hspace{1 cm}t\geq 0 \\
0\hspace{1 cm}& \hspace{1 cm}t<0\\ 
\end{split}\right.
\label{eq:xnsignal}
\end{equation}
where $A_n\,$(V/s$^m$), is the amplitude of the biological signal, $B_n\,$(s$^{-1}$), is the decay time, and $m_n$ is temporal consistency dependence. The function decays to zero within about $10/B_n\,$(s) and has a peak at $t = m_n/B_n\,$(s). 
These parameters depend on the type of cell analyzed (e.g. this model can describe both neuron or muscle cells). For our model it is important to know that the characteristics of the nervous cells slightly change in terms of amplitude, velocity, and firing rate. For this reason each neuron can have a different shape.

It is worth noting that a spike can also be represented in the frequency domain. This is obtained by considering the Fourier transform of (\ref{eq:xnsignal}), which gives
\begin{equation}
    X_n(f)=\frac{m_n!A_n}{(B_n+j2 \pi f)^{m_n+1}}.
\end{equation}
The spectral characterization provides an invaluable information for the development of the unsupervised hybrid kernel classification approaches that are investigated in Sec. \ref{sec:ann}.

However, the ENG is not given just a single spike, but it is the sum of several pulses. The frequency of spikes is called firing rate and determines the type of motor/sensory information a neuron sends to other cells. The firing rate depends on the stimulus, the ion channels, and the neuron's properties. It varies widely and reflects neural activity and coding.
The interspike interval over time is typically modeled as a Poisson process~\cite{richardson2000modelling,ostojic2011interspike}. The stochastic and pulsatile nature of spikes provides a coding mechanism that allows the nervous system to transmit information.
Accordingly, the proposed noiseless ENG signal observed by the $k$th electrode, with $k=1,\ldots, K$, can be modelled as
\begin{equation}  u_k(t)\hspace{-.07cm}=\hspace{-.1cm}\sum_{n=1}^{N} \hspace{-.07cm}h_{k,n}\hspace{-.1cm}\left(\hspace{-.1cm}\sum_{i=1}^{I_n} \hspace{-.05cm}x_{i,n} (-t \hspace{-.02cm} + \hspace{-.02cm}  T_{i,n}^{aff})\hspace{-.1cm} +\hspace{-.15cm}\sum_{j=1}^{J_n} \hspace{-.15cm} x_{j,n}(t \hspace{-.02cm} - \hspace{-.02cm} T_{j,n}^{eff})\hspace{-.1cm}\right)\hspace{-.1cm},
   \label{eq:uk}
\end{equation}
where $I_n$ is number of spikes in the $n$th afferent axon, $J_n$ that in the efferent one, and $T_{i,n}^{aff}$ ($T_{j,n}^{eff}$) indicates the instant of time corresponding to the $i$th ($j$th) pulse in the afferent (efferent) direction. 
The value of $h_{k,n}$ is the lead field given by \cite{Leadfield}
\begin{align} \label{eq:h_kn}
h_{k,n} = \frac{1}{4 \pi \sigma} \left(-\frac{1}{||\mathbf{n} - \mathbf{k}||^2}\right),
\end{align}
where $\sigma$ represents the temporal-dependent conductivity, $|| \cdot ||$ is the Euclidean norm, and $\mathbf{n}$ and $\mathbf{k}$ are the $3$-dimensional column vectors defining the position of the axon and of the electrode, respectively, with respect to an arbitrary reference system. $h_{k,n}$ is influenced by the temporal conductivity and the spatial separation between the $n$th source and the $k$th electrode positions
\footnote{It is worth noting that, there is a slow temporal dependency of $h_{k,n}$ over days/weeks. In fact, an inflammatory response can occur due to the presence of the cuff electrode. This leads to the formation of a fibrotic capsule that generates a material with insulating properties, thus reducing the measured values up to $30$\% over time before stabilizing after a few months~\cite{optogenetica}.}, Fig. \ref{fig2b}.

The spikes and their associated delays are generated according to a Poisson distribution once the firing rate in each of the two directions is defined. Indeed, the proposed ENG MIMO model accounts for axons with bidirectional signal propagation speeds, afferent and efferent. Note that, although in (\ref{eq:uk}) the reported number of axons is $N$, they are not all active at the same time, depending on the type of stimulus (sensory, referred to $i$, and motor, referred to $j$) and its intensity. To take this aspect into account we set to zero the corresponding $h_{k,n}$ for the non-active $n$th axon.

The overall noiseless signal observed by the the cuff electrode with $K$ contacts can be represented in matricial form as 
\begin{equation}
\mathbf{u}(t)=\mathbf{H} \cdot \mathbf{x}(t),
\label{eq:MIMOModel}
\end{equation}
where $\mathbf{H}$ is the $K \times N$ lead field matrix \cite{Leadfield}, whose entries are defined in \eqref{eq:h_kn} and  $\mathbf{x}(t)$=$[{x_1(t),\ldots, x_N(t)}]^T$. 

However, in order to define a realistic model, we must take into account that besides the noiseless ENG signal defined in (\ref{eq:uk}), the $k$th electrode is also impaired by the presence of distortion components. The two most significant ones are the interference generated by the electromyographic signal $\mathrm{emg}(t)$, which originates from muscles and is the main antagonist of the ENG signals since it lies in the same bandwidth, and by the thermal noise $w_k(t)$, which is white Gaussian with zero mean and unknown power $\sigma_k^2$. While the thermal noise component over the $K$ electrodes are all independent, the $\mathrm{emg}(t)$ signal, originated externally to the cuff, can be considered constant and uniform over the entire cuff~\cite{EMG_uniforme}. Note that, the electromyogram signal has a characterization similar to that of the ENG signal. However, after its analog filtering done in the PNI the $\mathrm{emg}(t)$ signal follows a zero-mean Gaussian distribution with unknown variance~\cite{EMG_gaussian}.
With the introduction of these two distortions, the measured signal at the $K$ multi-contacts cuff in matricial form turns out to be
\begin{equation}
\mathbf{y}(t) = \mathbf{u}(t)+\mathbf{d}(t),
\label{eq:y_k}
\end{equation}
where $\mathbf{y}(t)=[y_1(t),\ldots, y_K(t)]^T$ and $\mathbf{d}(t)=[d_1(t),\ldots,$ $ d_K(t)]^T$ is the distortion vector, whose $k$th entry is
$d_k(t) = \mathrm{emg}(t)+w_k(t)$.

The introduced model allows for the generation of a synthetic ENG signal and its measure with a multi contact cuff electrode. Its main characteristic is the ability to reproduce the behavior of an ENG signal where the spatial diversity of the spikes propagating in the axons is preserved~\cite{fabiana}. This feature makes it possible to evaluate the performance of methods able to extract the spatial-temporal characteristics of the ENG signal. 

Indeed, the ANN-based classification techniques investigated in Sec.~\ref{sec:classification} exactly aim to extract such spatial-temporal features, recognizing the presence of spikes and their firing rate through the use of kernels in the classifiers. For this reason the model defined by (\ref{eq:MIMOModel}), (\ref{eq:uk}), and (\ref{eq:y_k}) is crucial for defining the parameters associated with the kernels such as their length and number. This in-depth spike analysis provides crucial information on neural connections and stimuli. In fact, the presence of a spike indicates that a connection is occurring, while the spike density frequency $\lambda_n(t)$ over time represents the intensity. Our model considers spikes as central elements for classification, allowing APs in ENG nerve signals to be effectively identified and distinguished using kernels.

\subsection{ENG modelling as a networked system}

The nervous system acts as a nanoscale networked system that transmits signals between the brain and the periphery and vice versa. Understanding the propagation of this information within the body is of fundamental importance, especially if we consider the numerous axons contained in the nerves. This field of research is of crucial importance in areas of nanotechnology, communication at the molecular level, and medicine. 
In our study, the introduced ENG MIMO model offers the possibility to analyze how the signals produced by different axons interact. These interactions form the ENG signal measured by the cuff electrode that is wrapped around the nerve. Moreover, it allows numerical simulations via multiphysics software like, for example, COMSOL~\cite{fabiana}, capable of generating realistic synthetic data for meaningful comparisons in terms of amplitude and intrinsic characteristics of the signal. Through simulations it is possible to explore the propagation mechanisms inside the nerve. This offers a complete view of a neural electromagnetic nanonetwork where the complex and aggregate behavior of numerous axons is characterized to describe the emission of a more complex signal like, for example, the ENG one here considered .

\subsection{ENG signal acquisition}

To gain insights on the main features of the measured ENG signals, i.e. amplitudes, spectral content, and impairments, we consider as a reference of implanted device for signal acquisition Senseback, which is described in~\cite{[9]Senseback}. Senseback is a bidirectional wireless post-implant reprogrammable PNI designed to enable chronic peripheral electrophysiology experiments, potentially longer than $6$ months. It consists of a fully semi-flexible PNI with up to $32$ channels per cuff electrode that is capable of measuring and stimulating the nerve at multiple points and of bidirectional data transferring with the external PU via BLE communication protocol. In terms of average power consumption, Senseback operates in the range $20-100\,$mW, depending on the signal sampling mode. From the point of view of bandwidth occupation, ENG signals are in the range between $500\,$Hz and $7\,$kHz with a energy peak at about $2\,$kHz~\cite{ENGbanda}. Note that, the lowest and the highest frequency components dependent from the type of electrode~\cite{emg2}. For this reason, the maximum sampling rate of SenseBack is $20\,$kHz for each channel.  

The sampled measured ENG signals are characterized by weak amplitudes with an effective value of about $50\,$$\mu$V~\cite{ENGbanda}. 
However, this amplitude can undergo to variations in its strength due to incorrect placement of the cuff electrode ~\cite{silveira2018influence}. Moreover, it is affected by different types of distortions, the most important discussed in what follows~\cite{Demosthenous}:
\begin{itemize}
    \item The interference generated by the EMG signal associated with the muscular activity on the same frequency band of the ENG. The EMG is stronger than the ENG one. Its effective value is around $5\,$mV~\cite{Rieger} and its frequency band also extends from $0$ to $10\,$kHz. However, about $95$\% of the EMG energy is below $800\,$Hz~\cite{emg1}-\cite{emg3}. The distortion introduced by the EMG can be reduced by applying a highly selective pass band filter.
    \item The effect of the thermal noise $w_k(t)$, which is white on the entire frequency spectrum.
    \item Main power line interference at $50\,$Hz. Its effect can be attenuated by means of a notch filter~\cite{[7]Federica}.
\end{itemize}
Considering that most of the spectral content of an ENG signal is below $2.5\,$kHz~\cite{[6]Elisa, [7]Federica}, the contribution of all the above mentioned impairments can be mitigated in the digital pre-processing by applying a band pass filter with cut-off frequencies between $0.8$ and $2.5\,$kHz to the received samples of ENG signal.

The pre-processing unit of Senseback implements multiple parallel signal conditioning chains, one for each recorded channel, that include amplification, overall gain up to $50$$\,-\,$$70\,$dB, and analog filtering steps. 
In addition, the device also contains a flash memory of $512\,$KB and a RAM of $64\,$KB that eases the data acquisition. It is also equipped with a blanking circuit that enables simultaneous and timely signal detection along with nerve stimulation. This feature helps to mitigate acquisition disturbances for more effective and reliable performance~\cite{[9]Senseback}.

\section{Real-time ANN-based ENG classification}
\label{sec:classification}

In this section we first highlight the data set and the pre-processing we applied to validate out system model. Then, we present an analysis of the latency introduced by the sequence of operations implemented by the ND\&S system. Finally, we explore different ANN architectures that we used to extract the different sensory stimuli contained in the analyzed recorded data set.
\begin{figure}[!t]
\centering
\includegraphics[width=0.8\columnwidth]{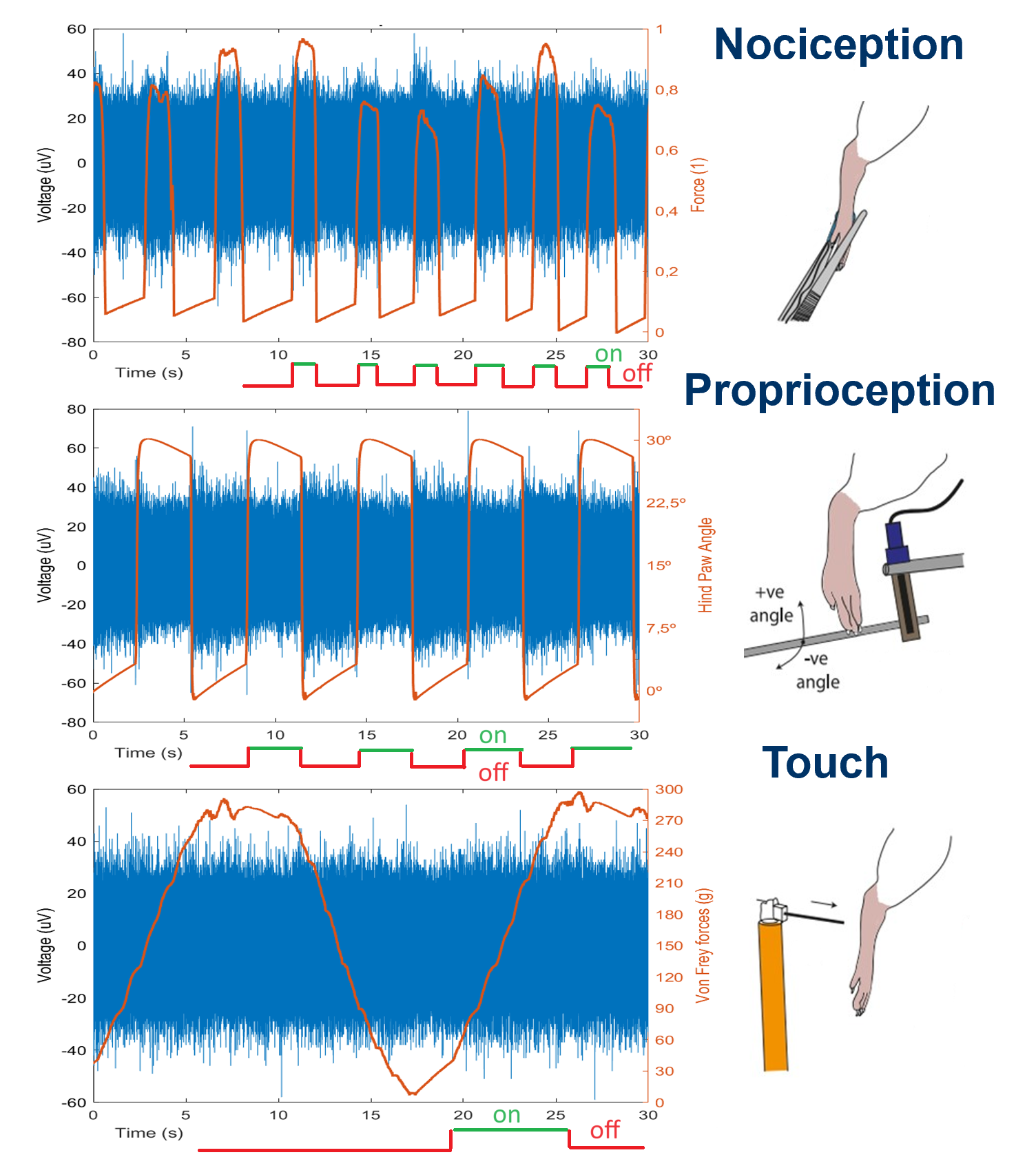}
\caption{Representation of the dataset studied and grouping of the $4$ classes for classification \cite{[12]Identification}.}
\label{fig:stimuli_class}
\end{figure}

\subsection{The Data Set and its Pre-processing} \label{sec:preprocessing}

\begin{figure*}[!t]
\centering
\includegraphics[width=2\columnwidth]{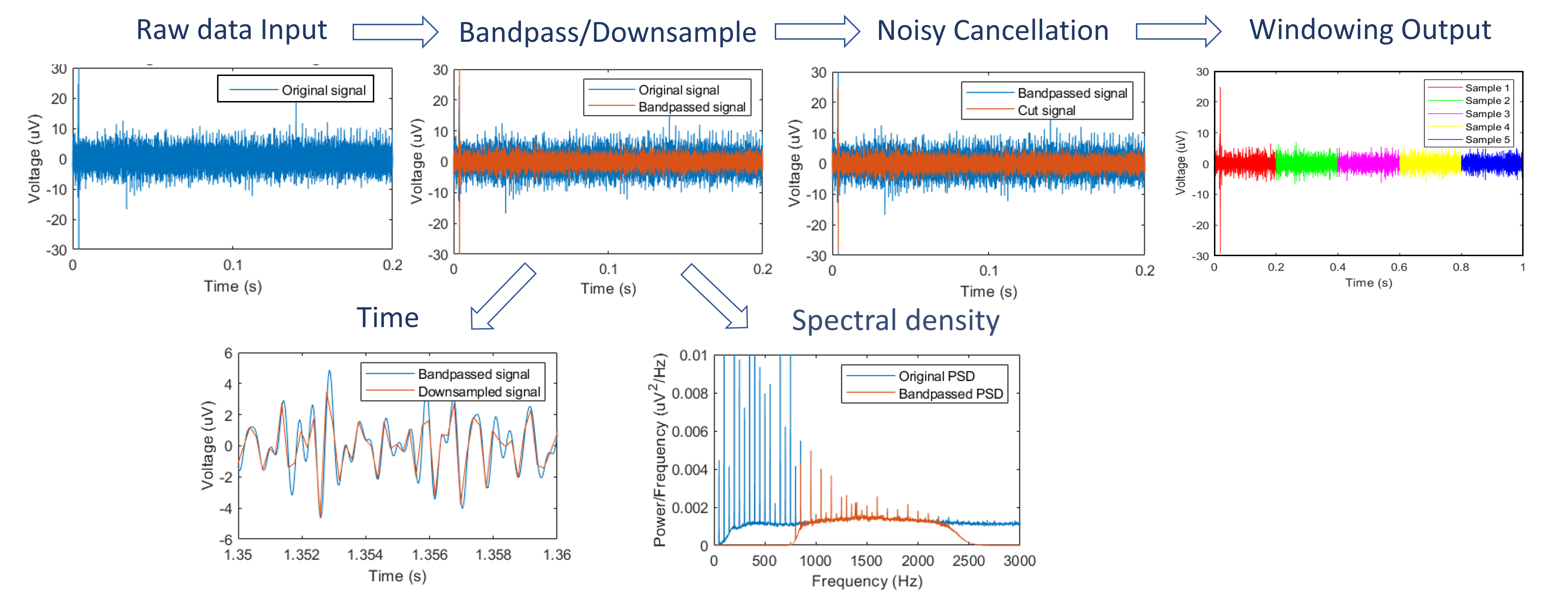}
\caption{From the raw ENG data set to processed.}
\label{fig:flux}
\end{figure*}

Our classification was tested on real ENG signals measured from the sciatic nerve of three rats. The used data set is described in \cite{ [12]Identification}. It contains nerve recordings of mechanically stimulated sensory nerve activity of rats. The classification is performed for $4$ distinct types of activities: dorsiflexion, plantaflection, touch, and pain, as exempliefied in Fig.~\ref{fig:stimuli_class}. The stimuli were applied with an automated system to limit the variability between the measurements and eliminate the influence of the human operator. The automated system works by alternating stimulation and rest periods of $3\,$s each, in case of pain of $1\,$s each. The ENG signals were recorded at sampling frequency of $30\,$kHz using a cuff electrode with $16$ read channels arranged in a matrix configuration of $4$ rings, each having $4$ contacts. 

Since the ultimate goal is to provide real-time prediction, the pre-processing phase is kept as simple as possible and robust classifiers were chosen to handle the noisy data. The following pre-processing steps, whose results are reported in Fig. \ref{fig:flux} were implemented
\begin{enumerate}
    \item \textbf{Bandpass filtering}: The signals undergo to an $8$th order Butterworth bandpass filter in the range $0.8-2.5\,$kHz to remove the EMG's low-frequency contribution and other types of high-frequency noise;
    \item \textbf{Downsampling}: The signals are downsampled at $5\,$kHz to reduce computational load and to preserve the main features for signal classification;
    \item \textbf{Thresholding}: Due to the high voltage peaks present in the raw data signals, which can be classified as outliers for ENG signals, an empirical  tresholding operation is performed. The physiological threshold is set to $30\,\mu$V so that measurements whose absolute value exceeded this value were canceled. This threshold cleans the data at the cost of eliminating a very small portion of them;
    \item \textbf{Signal windowing}: The signal is divided in non overlapping windows of different durations $T_w$. The duration of the windows must be chosen to make possible the real-time operations of the investigated classifiers.
\end{enumerate}

\subsection{Analysis of the latency for real-time classification} \label{sec:delays}
Since our goal is to implement real-time classification, it is important to quantify the overall delay introduced by the different operations implemented by an ND\&S system, as described in Sec.~\ref{sec:preprocessing}. In the following we evaluate the different times that contribute to the overall feedforward time $T_f$ in \eqref{eq:Tf}. To do this we refer again Senseback, which can be considered as the state-of-the art of PNI devices. With reference to Senseback technical specifications, its clock frequency is equal to $16\,$MHz~\cite{[9]Senseback}. This results in a value of $T_a$ in (\ref{eq:Tf}) in the order of nanoseconds, which can be therefore neglected in the evaluation of the overall feedforward time $T_f$. Considering a cuff electrode with $K = 16$ contacts and an ADC with $b=10\,$bit resolution, the sampling at frequency of $f_s = 5\,$kHz~\cite{[6]Elisa,[7]Federica} gives a payload of $P=800\,$kbit in a window of $T_w=1\,$s (\ref{eq:Tf}). Since the maximum payload data-rate of BLE is $R_{ble} = 1.4\,$Mbit/sec~\cite{[9]Senseback}, the minimum uplink transmission time $T_u$ in (2) can be easily calculated as 
\begin{equation} 
\label{eq:tu}
    T_u = \frac{P}{R_{ble}}\,\,\,[\text{s}].
\end{equation}
For $T_w=1\,$s we obtain $T_u\approx 572\,$ms. Note that, the value of $T_u$ changes proportionally to $T_w$, and it goes from a value of $T_u\approx 29\,$ms when $T_w=50\,$ms to a value of $T_u\approx286\,$ms when $T_w=500\,$ms. The choice of $T_w$ is one of the most critical aspects in the design of ND\&S systems since its value directly affects the residual time $T_c$ that remains to implement classification given the limit defined on the feedforward time $T_f\leq 300\,$ms by the human perception delay. Note that, as discussed in~\cite{anna}, the value of $T_u$ could be reduced by applying source compression. About the downlink transmission time $T_d$, it is approximately equal to $2\,$ms. This value can be obtained by considering a payload data packet of $244\,$bytes. Finally, the value of $T_s$ is approximately equal to $20\,$ms, as reported in~\cite{stimuly}. 

We remark here that the above evaluation of the delays introduced by the different operations implemented by a ND\&S system allows us to obtain the time $T_c$ available for the classification. In the next section we discuss the ANN-based classification architectures proposed in this paper.

\subsection{Exploiting the ENG MIMO model in the unsupervised hybrid kernel classification approach} \label{sec:collegamento}
One of the main functions of the ENG MIMO model introduced in Sec.~\ref{sec:model} consists in the extraction of spikes's features from the measures taken by the multi-contact cuff electrode. Such an operation allows to obtain the model's parameters that can be exploited to bear the practical implementation for the classification of ENG signal by means of the investigated ANNs. The procedure to achieve this is explained in what follows.

The proposed model is able to reflect the physiological behavior associated with the human axon-based propagation of the information. This is enclosed in~\eqref{eq:uk}, which highlights the main parameters of interest $T_{i,n}^{aff}$, $T_{j,n}^{eff}$, $x_{i,n}(t)$, $x_{j,n}(t)$, and $h_{k,n}$. The same association can be exploited to define the size of the kernels to be used in the considered ANNs. To do this, an action potential is plotted using the model in \eqref{eq:xnsignal}. Then, the parameters $A_n$, $B_n$, and $m_n$ are tuned to establish the best match between what measured and the characteristics of the spike generate by the $n$th axon, i.e. its total duration and that of its peak, and finally define the kernel dimension. We consider the total duration to be sure that the kernel is able to capture all the components of interest after the training phase.

From the available data sets we observed that the duration of the action potential is $3\,$ms, while the peak has a duration comprised between $0.5\,$ms and $1\,$ms. Consequently, the spectral content, i.e. the inverse of the peak duration, is in the range $1\,$-$2\,$kHz. With reference to the spike duration, if it lasts $3\,$ms in total, and the $f_s =5\,$kHz, then the number of spike samples is $15$. In order to find the optimal kernel length we iterated by first training the ANNs with $15$ samples and then adjusted this number by considering the neighboring values. This adjustment is required due to the physiological fluctuation that characterize different subjects. For CNN and LSTM we obtained an optimal value of $9$, while for IT we found $14$. For ENGNet, in place of considering the single spike, we increased the window dimension to $100$ samples to highlight the firing rate. This affects $T_{i,n}^{aff}$ and $T_{j,n}^{eff}$. For the spatial dimension, we defined another kernel with size equal to the number of channels $K$, accounting for the correlation between the source position and $h_{k,n}$. 
\begin{figure*}[!t]
\centering
\includegraphics[width=1.9\columnwidth]{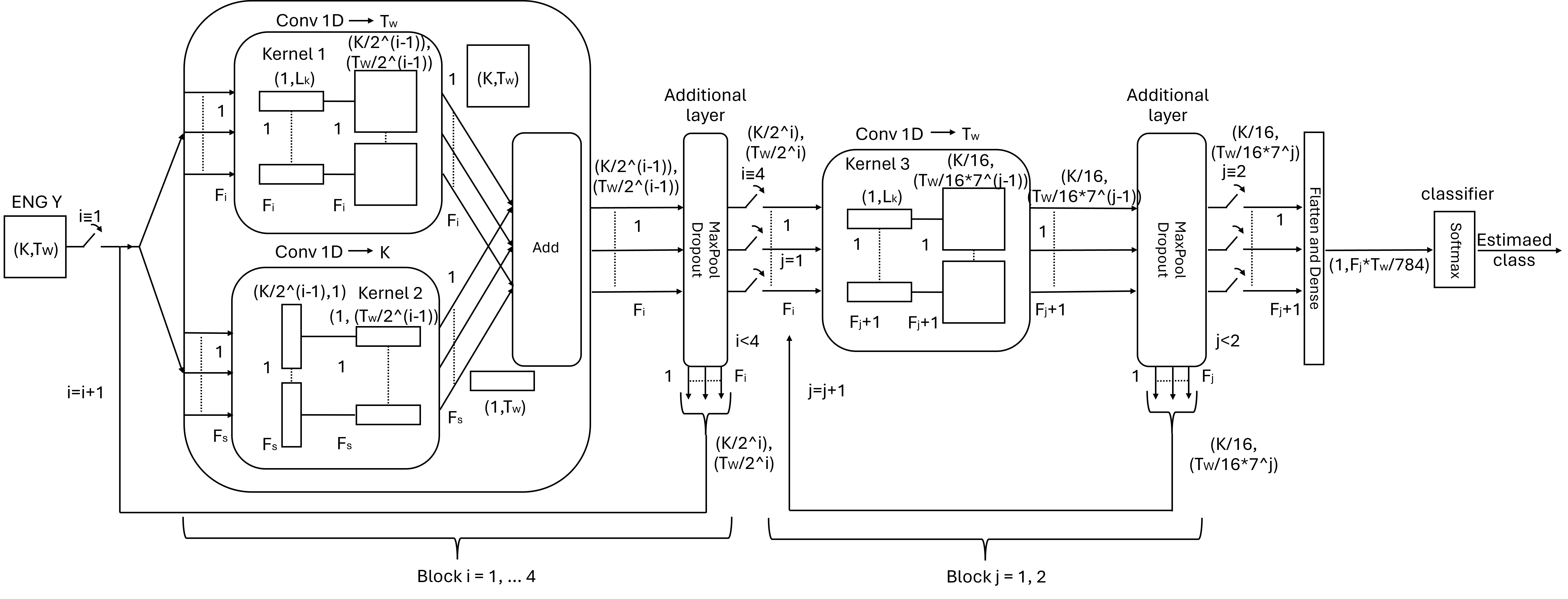}
\caption{ Schematic representation of the baseline CNN architecture. $F_i$ and $F_j$ are the number of the linear temporal kernel, and $F_s$ of the spatial kernel, applied for each block, equal to 32 each, $L_k$ is the temporal length of the kernel, $K$ in the number of electrodes, $T_w$ is the window length. $\mathbf{Y}$ is the input of the model while there are $4$ possible output classes. The rectangles represent column or row vectors, while squares represent matrices. The repeated blocks are represented by the index $i$ and $j$ with a feedback arrow. In those cases the output of the block become the input itself.}
\label{fig:CNN}
\end{figure*}

The ANN-based classification phase employs kernels that are selected to emphasize specific frequency characteristics that are closely linked to the ENG signals. The fundamental connection between the model and the dataset guides parameter choices and kernel configurations based on obtained signals, facilitating the effective identification and distinction of AP and their recurrence over time in nervous signals.

\subsection{Artificial Neural Networks}
\label{sec:ann}

We investigate four different types of architecture with different complexity. Each of them is adapted to take advantage from the proposed ENG signal modelling. 
Changes include the use of 1D kernel, kernel sizes, number of convolutional blocks, implementation of an early stopping algorithm, and other minor optimizations specific to each network. The use of 1D kernels, allows to optimize the computational efficiency, accelerating training and prediction times, making the network suitable for real-time applications. Variations in kernel sizes and other parameters were applied progressively through iterations in the network. Finally, the convolutional block was then repeated by adding blocks until a plateau in performance was reached. Furthermore, an early stopping algorithm has been implemented to choose the best performing training period. In this way the models used for the test is not the one at the end of training after a fixed number of epochs but could vary depending on the case. This led to better results while also reducing the inaccuracy associated with the forecast.

\subsection*{Convolutional Neural Networks (CNNs)} 

CNNs are used to emphasize meaningful spatio-temporal relationships that underlie complex multivariate time-series data. 
Their application to the classification of 1D biological signals has been already considered in previous works. In particular, in \cite{[14]koh2020selective} CNNs were used to classify the spatio-temporal signature of features extracted from the ENG signal, while in \cite{[15]luu2021deep} a CNN was used to decode motor intent (i.e., efferent ENG signal) on sliding windows of ENG data to look like online engine decoding. 
\begin{figure*}[!t]
\centering
\includegraphics[width=2\columnwidth]{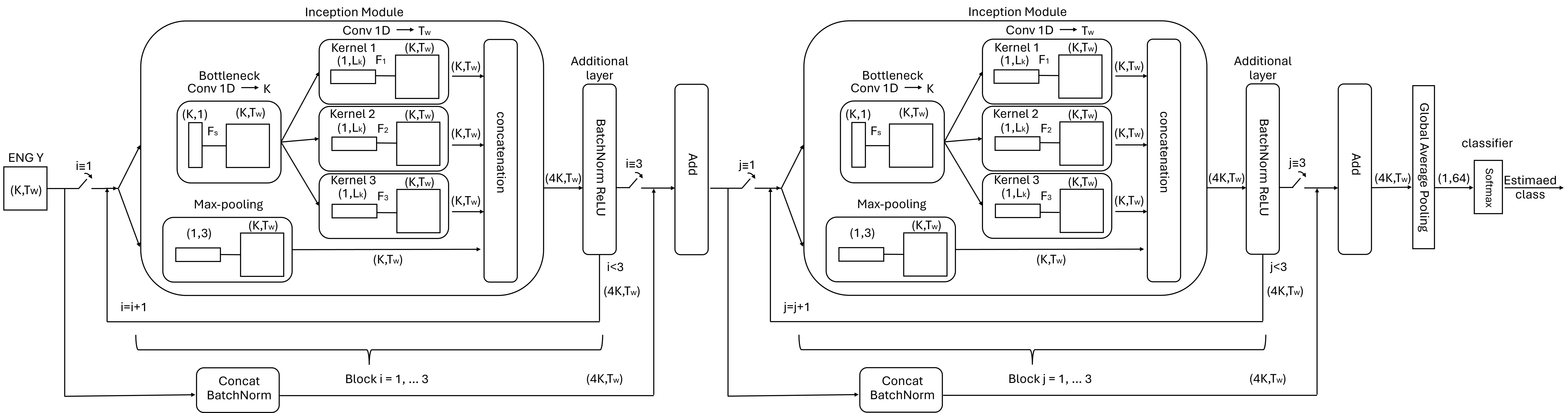}
\caption{Schematic representation of the baseline IT architecture. $F_1$, $F_2$ and $F_3$ are the number of the linear temporal kernels, and $F_s$ of the spatial kernel, applied for each block, equal to 1 each, $L_k$ is the temporal length of the kernel, $K$ in the number of electrodes, $T_w$ is the window length. $\mathbf{Y}$ represents the input of the model while there are $4$ possible output classes. The rectangles represent column or row vectors, while squares represent matrices. The repeated blocks are represented by the index $i$ and $j$ with a feedback arrow. In those cases the output of the block become the input itself.}
\label{fig:CNNKernel}
\end{figure*}
\begin{figure*}[!t]
\centering
\includegraphics[width=2\columnwidth]{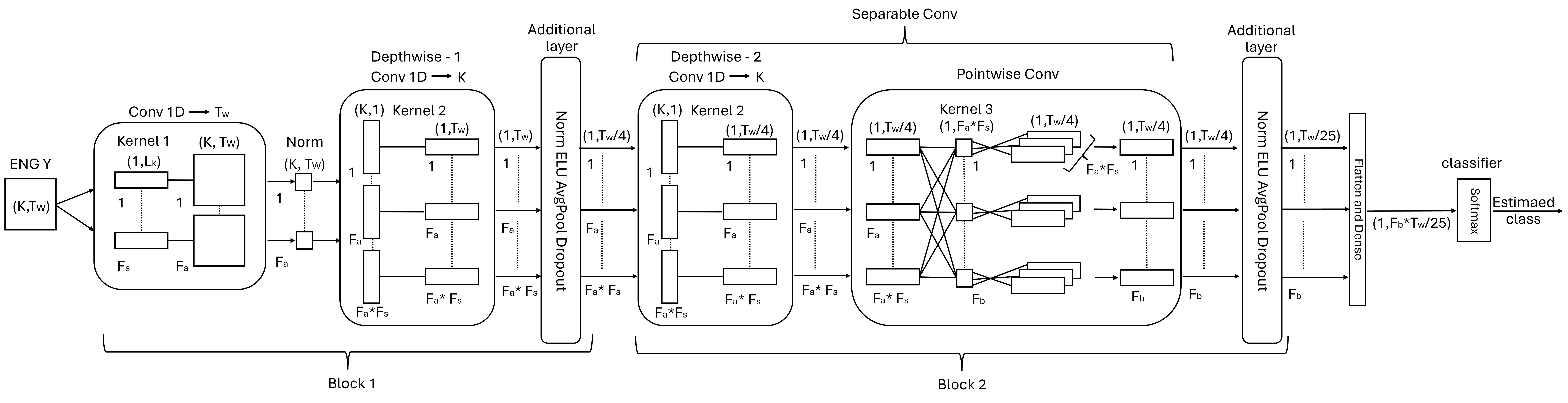}
\caption{Schematic representation of the baseline ENGNet architecture. $F_a$ and $F_b$ are the number of the linear temporal kernel, and $F_s$ of the spatial kernel, applied for each block, equal to 16, 32, 2 respectively, $L_k$ is the temporal length of the kernel, $K$ in the number of electrodes, $T_w$ is the window length. $\mathbf{Y}$ is the input of the model while there are $4$ possible output classes. The rectangles represent column or row vectors, while squares represent matrices.}
\label{fig:ENGNet}
\end{figure*}

In our study, a CNN was built and trained to serve as a gold standard against which to compare the other ones. The proposed CNN consists of four main layers: i) \textit{Convolutional layer.}, ii) \textit{Dropout layer},  iii) \textit{Pooling layer}, and iv) \textit{Fully Connected Layer}. Its architecture, which has been adapted from \cite{[7]Federica}, is shown Fig. \ref{fig:CNN}. The main difference with respect to~\cite{[7]Federica} consists in  the addition of an early stopping algorithm.

Since the processed signals are 1D, a combination of two 1D convolutional layers is applied. The information of all the electrodes is compressed into the same time step by using two kinds of 1D kernels, one for each dimension: the temporal kernel and the spatial kernel. The temporal one has dimensions $D\times L_k$, where $D$ represents the 2D dimensionality and $L_k$ is the temporal kernel length.  $L_k$ is equal to $9$ samples, emphasizing the spike features as discussed in Sec. \ref{sec:collegamento}. The spatial kernel has $K\times D$ dimension, where $K$ is the total number of rows of the input, that in our case is equal to $16$ electrodes. Since we are working with 1D kernel, we set $D$=1. 

The general 2D convolution applied is given by
\begin{equation}
    g(m,n)=\sum_i \sum_j F(i,j)Y(m-i,n-j),
    \label{eq:2DconvKernel}
\end{equation}
where $Y(m,n)$ is the ENG signal $Y$ reported in Sec. \ref{sec:model}, $F(m,n)$ is our kernel. Since we have a variable number of kernels in the convolutional blocks reported in Figs.~\ref{fig:CNN}-\ref{fig:ENGNet}, 
a different subscript is used\footnote{$F_s$ represents the number of spatial kernel, and $F_i, F_j, F_1, F_2, F_3, F_a, F_b$ represent the number of temporal ones.}. 
The convolution is performed along the $m$th and the $n$th row and column of step $i$th or $j$th.
$Y$ is a matrix with dimension $K\times T_w$, with $T_w$ equal to the time window chosen, and it represent the input of the model. The convolution in (\ref{eq:2DconvKernel}) is used for all the considered neural networks in the following. 
\begin{figure}[!t]
\centering
\includegraphics[width=1\columnwidth]{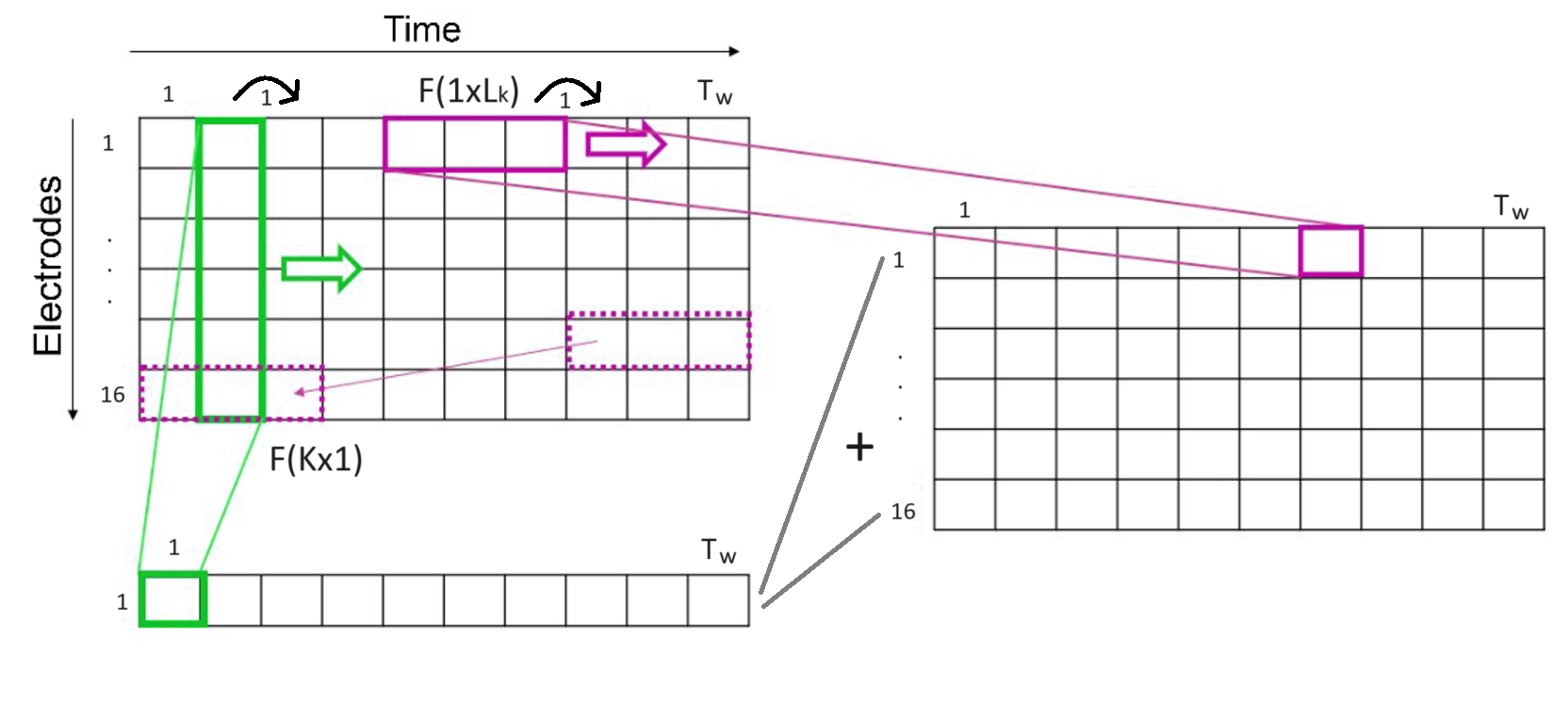}
\caption{Representation of the 1D convolutional operation.}
\label{fig4}
\end{figure}

The kernels perform a point wise convolution on the input moving through the entire input length. Both kernels are moved with a step of $1$ sample and same padding, maintaining the size fixed throughout the convolution process, as shown in Fig.~\ref{fig4}. This creates two feature maps, on for the temporal and one for the spatial convolution, that are added element wise and passed down to the next layers. After this calculation, an activation function is applied that consists of a rectified linear unit (ReLU) applied to the output of the convolution block. The network is dropout and max pooled after each convolution level to perform prediction regularization and dimensionality reduction, respectively, as noted above. This architecture is iterated four times for the first convolutional block. Then a second convolutional block is repeated two times. This contains just a temporal kernel. The output of each convolution block is then used as input to the next one (as figured out by the feedback arrows reported in
the Fig. \ref{fig:CNN}). Then, the final output is applied to a fully connected softmax layer. The softmax layer transforms the feature maps into prediction probabilities and generates the network result. The network is trained for $50$ epochs using the backpropagation algorithm and it is shutdown using the early stopping algorithm added to the training phase to avoid further overfitting. Lastly, the best performing epoch in the validation data set is selected for the final prediction. 

\subsection*{Inception Time (IT)} 

The variant of the CNN proposed in~\cite{[16]ismail2020inceptiontime} is also investigated in this study. The architecture consists of building blocks similar to the above highlighted CNN. The arrangement of these layers is changed to emphasize the different characteristics of the signal as discussed in Sez. \ref{sec:collegamento}. The convolutional layer is structured using an \textit{inception module}, which is initially introduced for comprehensive image classification purposes~\cite{[17]szegedy2015going}. 
The architecture of this module is illustrated in Fig.~\ref{fig:CNNKernel}. It involves three main operations: i) \textit{Bottleneck layer} that compresses input multivariate time series (MTS) with dimensions $K \times T_w$, from $K$ electrode to $k$, where $k << K$. This significantly reduces computational overhead during forecasting. In this case we use a 1D spatial kernel with dimension $K\times 1$ to do the operation; ii) \textit{Bank of filters}, 1D temporal kernels with different lengths are applied in parallel to the output from the bottleneck layer. By processing the signal with different processing lengths, the network can highlight features present at different frequencies within the signal. Temporal kernel with lengths $L_k$ of $3$, $7$, and $14$ were selected for this application; iii) \textit{Max-pooling} that max-pools along the time dimension with length $T_w$. The sliding window is applied to the entire time series and its output is concatenated with the output of all kernels applied in the previous operation, generating the output of each inception module.

In our implementation, the IT network is trained for $50$ epochs and an early stopping mechanism is implemented to prevent overfitting. Both kernel size and early stopping hyperparameters have been determined through empirical tests conducted with available datasets. 
We keep the kernel lengths constant across different sample lengths of ENG signals. The network that is reported in Fig. \ref{fig:CNNKernel} is composed of $6$ Inception Modules which are grouped in $2$ blocks of $3$ Modules each. The second block's input receives an addition of a skip connection from the original input, just as its output receives an addition of its own input. 

Here, even if the inception module is applied multiple times, its temporal kernel size remains with constant lengths $L_k$ of $3$, $7$, and $14$. The output of each convolution in the Kernels layer and Max-Pooling layer is then concatenated and used as input to the next Inception module (the feedback arrows reported in Fig.~\ref{fig:CNNKernel}). A ReLU activation function is applied to the output of each the convolution block. Finally, the output is flattened and passed through a softmax layer to obtain network predictions.

\subsection*{Electroencephalogram Network (EEGNet)}

The Electroencephalogram Network (EEGNet) is another variant of CNNs specifically designed for Brain-Computer Interfaces (BCIs) applications introduced in~\cite{[20]lawhern2018eegnet, [23]feng2022efficient}.
EEGNet uses a combination of convolutional and clustering layers, leveraging depth and separable convolutions to efficiently process EEG data by capturing significant temporal and spatial patterns for classification. 
Its main advantage is that of being a compact architecture with fewer parameters than traditional CNNs. 
This makes EEGNet suitable for efficient and effective processing of electroencephalogram (EEG) signals, enabling better performance in BCI tasks while reducing computational complexity. In particular, its simplicity, the high computing power, and execution speed are suitable features in real-time applications.

The architecture of an EEGNet features two convolutional stages, using 2D convolutional filters to acquire EEG signals at different bandpass frequencies and depth convolutions to learn the spatial filters for each time filter. The approach also uses separable convolutions to reduce the number of parameters and combine feature maps efficiently. In the classification block, softmax classification is applied directly to extracted features without a dense layer, reducing the number of parameters in the model \cite{[31]tensorflow}. The EEGNet implementation includes the following layers:
\begin{itemize}
    \item Block 1 consists of a first convolutional pass that contains several 2D convolutional filters. The output produces a certain number of temporal feature maps exalting the ENG signal at different frequencies. Then, a depth convolution is used to create a spatial feature maps allows to choose the most performing channel recorded. In the EEGNet, these two operations allow to highlight frequency-specific characteristics and to have a better spatial channel selection, facilitating the information extraction of the ENG signals. 
    \item Block 2 based on separable convolution, which allows for a reduction in the number of parameters to fit and the explicit decoupling of the relationship within and between feature maps. Then, the extracted features are passed directly to a softmax classification level with a number of units that coincides with the number of classes in the data.
\end{itemize}

Additional layers are included in the EEGNet after convolutional steps in Block 1 and Block 2. Their main function is to apply batch normalization, ELU activation, average bundling, and abandonment techniques to improve feature learning and manage model complexity.

The Activation Function introduces nonlinearity into the model, allowing to learn complex patterns and representations in the data. The activation function is given by:  
\begin{equation}
    f(x)=    \left\{\begin{split}
x &  \hspace{1 cm}x\ge 1 \\
a(e^x-1)& \hspace{1 cm}\text{otherwise}\\ 
\end{split}\right.
\end{equation}
where x is Input tensor and	a is a scalar, slope of negative part, while A controls the value to which an ELU saturates for negative net inputs.

We adapted EEGNet to classify ENG signals.  
The main parameter changed in the EEGNet architecture to create an \textbf{electroneurographic network} (ENGNet) is the use of 1D kernel, the definition of its length and quantity to do the convolutional layers. This plays a crucial role in shaping the feature extraction capabilities of the network. By adjusting the kernel length, we can effectively capture the frequency information in ENG signals at different scales. 

The layers following the first are scaled according to the choice of maintaining the proportional settings proposed in \cite{[20]lawhern2018eegnet, [23]feng2022efficient}. A complete representation of ENGNet with all levels is given in Fig.~\ref{fig:ENGNet}. As in the others ANN, the convolution is split in two parts: the 1D temporal kernel with dimension $1\times L_k$, where $L_k$ represents the length of the kernel here fixed at $100$, and the 1D spatial kernel with dimension $K\times 1$, where $K\,$=16 is the number of electrodes. The Pooling layer reduces the spatial size of the feature maps by performing average pooling with a pool size of ($1, L_k/10$). 

In block 2, a first spatial kernel with size of ($K, 1$) performs the depth separable convolution, then a second temporal kernel with size of ($1,F_a*F_s$) performs the point wise convolution. This level effectively reduces the number of parameters while maintaining feature extraction capabilities. Then, the AveragePooling further reduces the spatial size using a pool size of ($1, L_k/25$).  

\subsection*{Long Short-Term Memory (LSTM) Network}
\begin{figure*}[!t]
\centering
\subfloat[]{\includegraphics[width=\columnwidth]{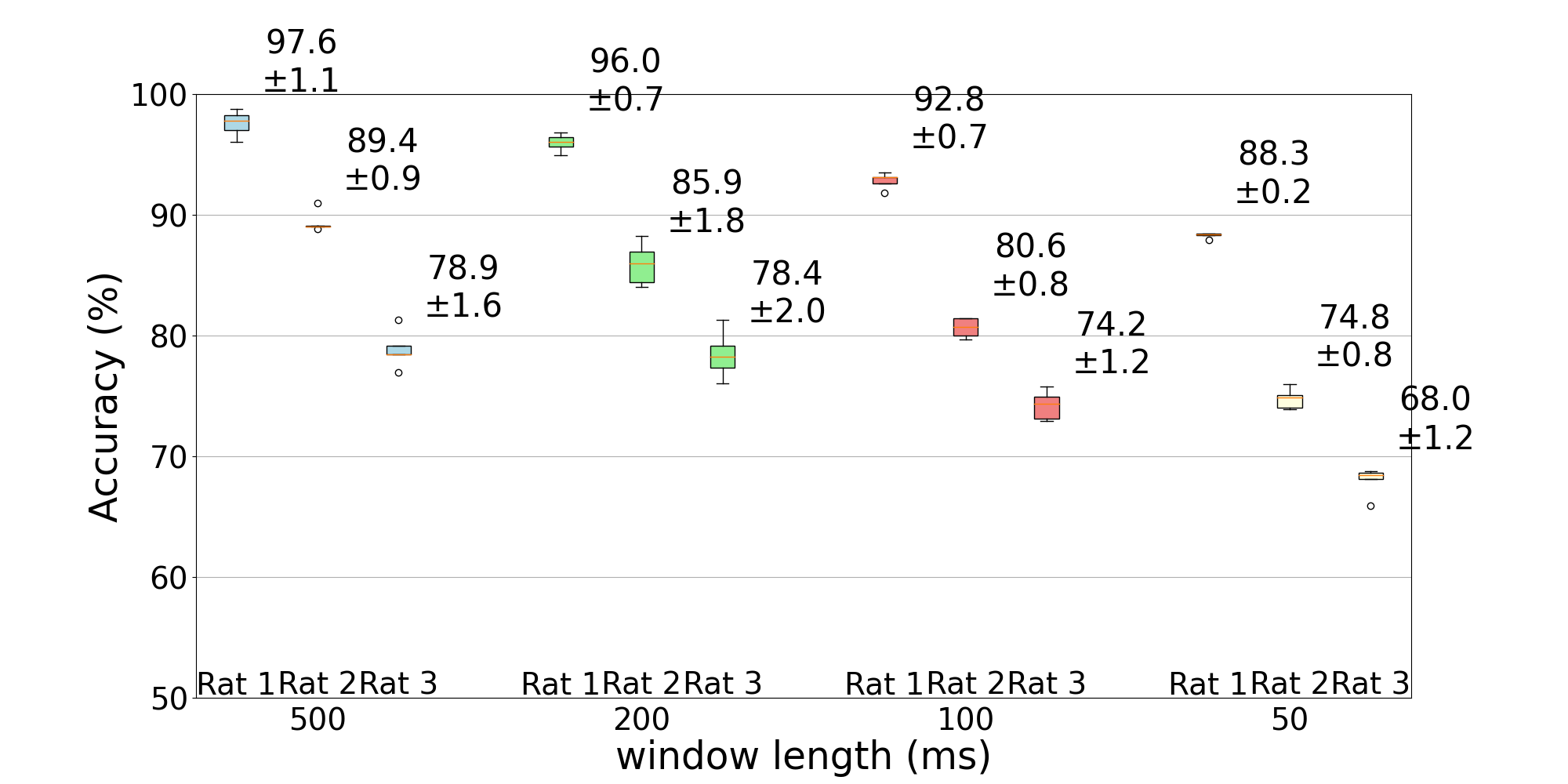}%
\label{fig_first_case}}
\hfil
\subfloat[]{\includegraphics[width=\columnwidth]{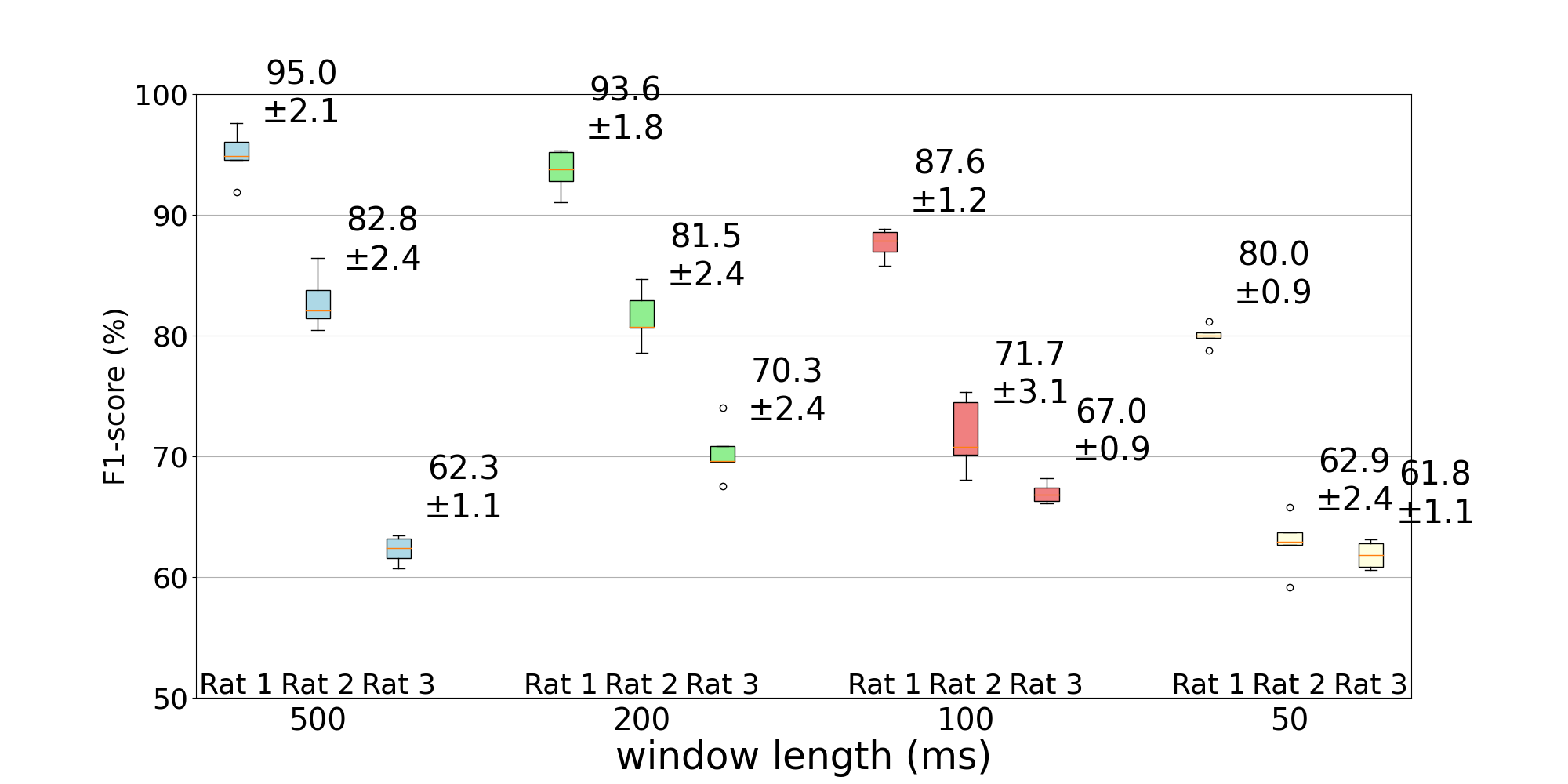}%
\label{fig_second_case}}
\caption{Accuracy (a) and F1-score (b) obtained by applying CNN and early stopping to the studied signals.}
\label{fig9}
\end{figure*}
\begin{figure*}[!t]
\centering
\subfloat[]{\includegraphics[width=\columnwidth]{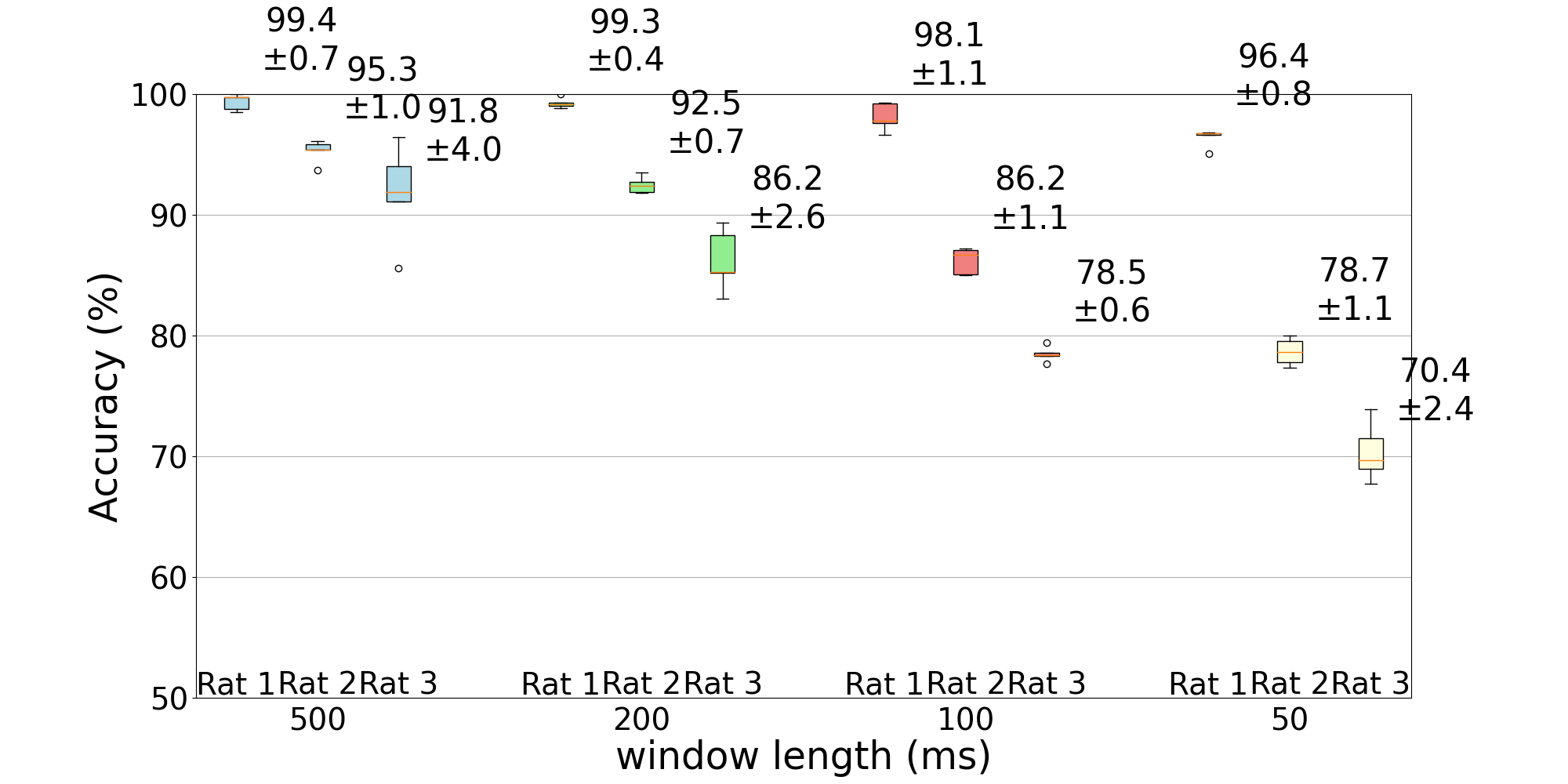}%
\label{fig_first_case}}
\hfil
\subfloat[]{\includegraphics[width=\columnwidth]{figure/Boxplot_CNN_F1.png}%
\label{fig_second_case}}
\caption{Accuracy (a) and F1-score (b) obtained by applying IT and early stopping to the studied signals.}
\label{fig10}
\end{figure*}
\begin{figure*}[!t]
\centering
\subfloat[]{\includegraphics[width=\columnwidth]{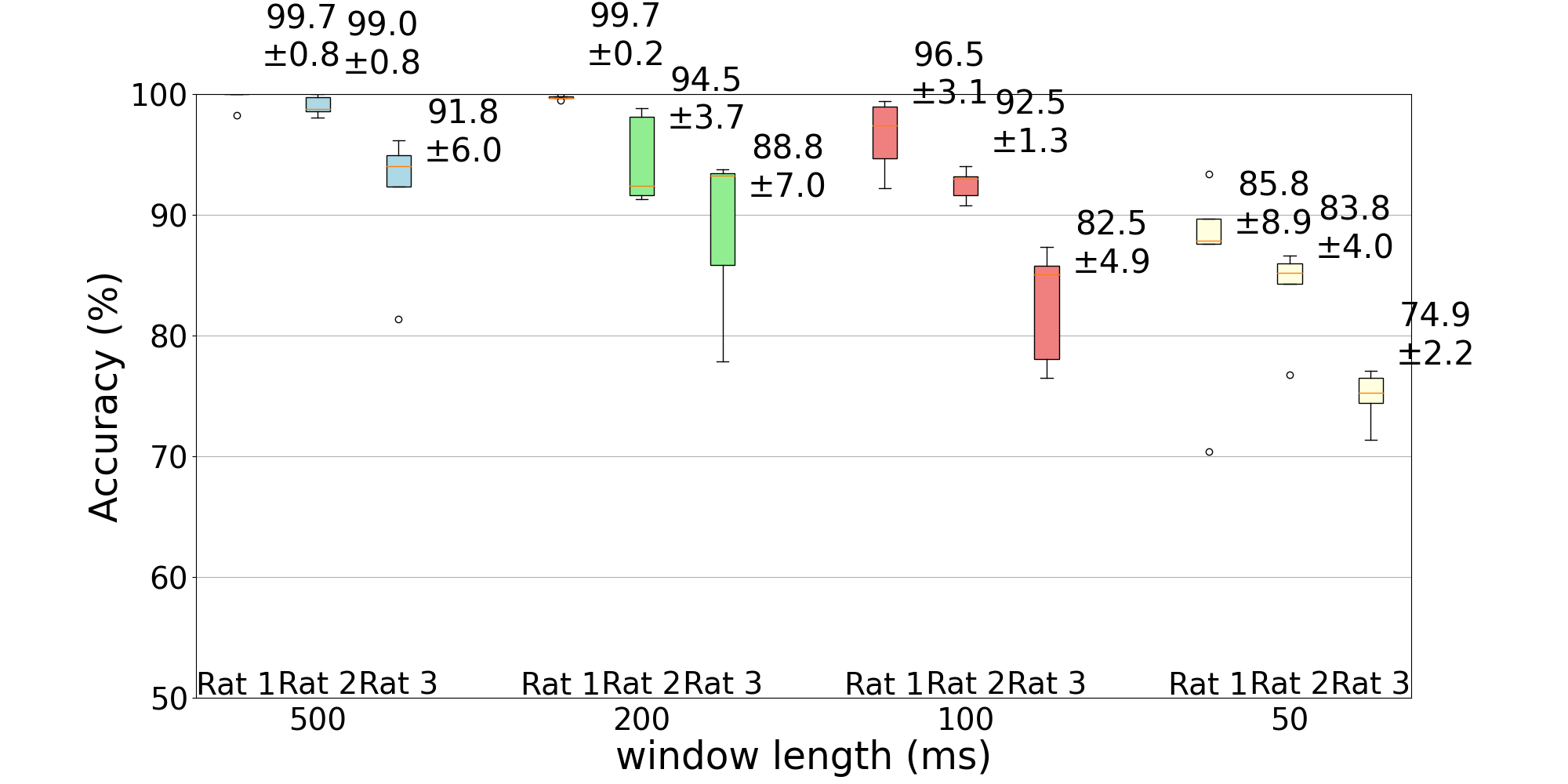}%
\label{fig_first_case}}
\hfil
\subfloat[]{\includegraphics[width=\columnwidth]{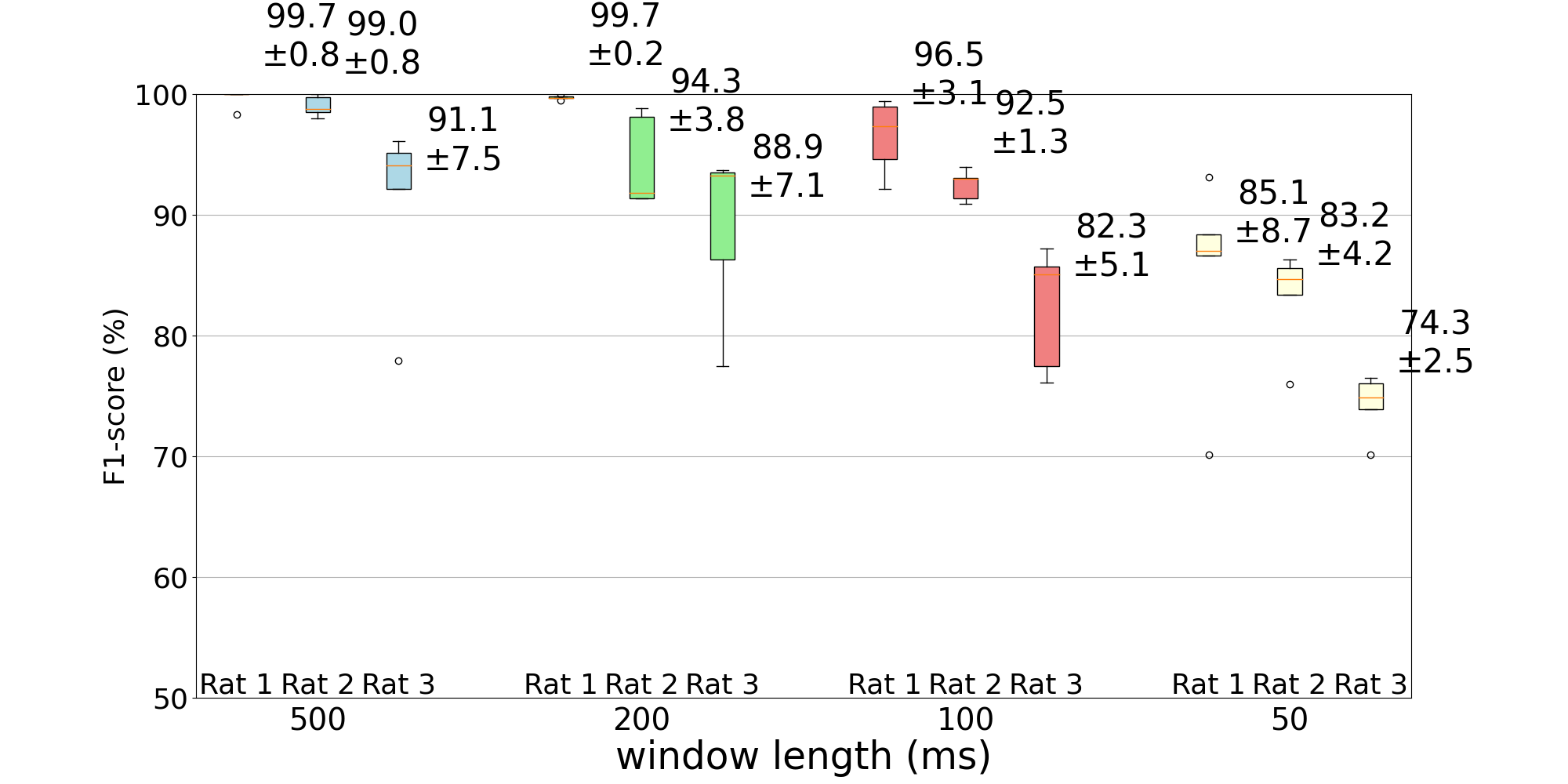}%
\label{fig_second_case}}
\caption{Accuracy (a) and F1-score (b) obtained by applying ENGNet and early stopping to the studied signals.}
\label{fig11}
\end{figure*}
\begin{figure*}[!t]
\centering
\subfloat[]{\includegraphics[width=\columnwidth]{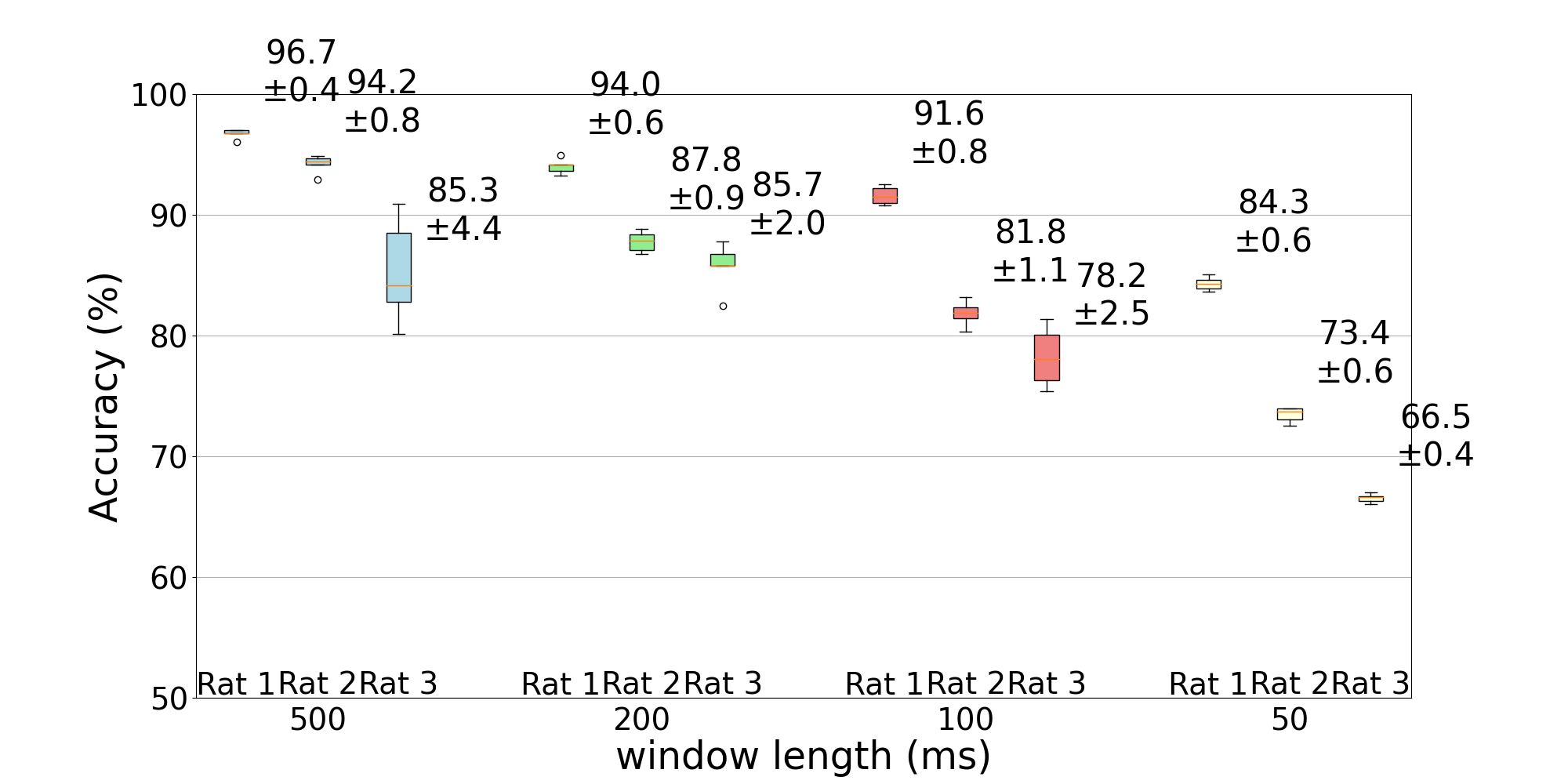}%
\label{fig_first_case}}
\hfil
\subfloat[]{\includegraphics[width=\columnwidth]{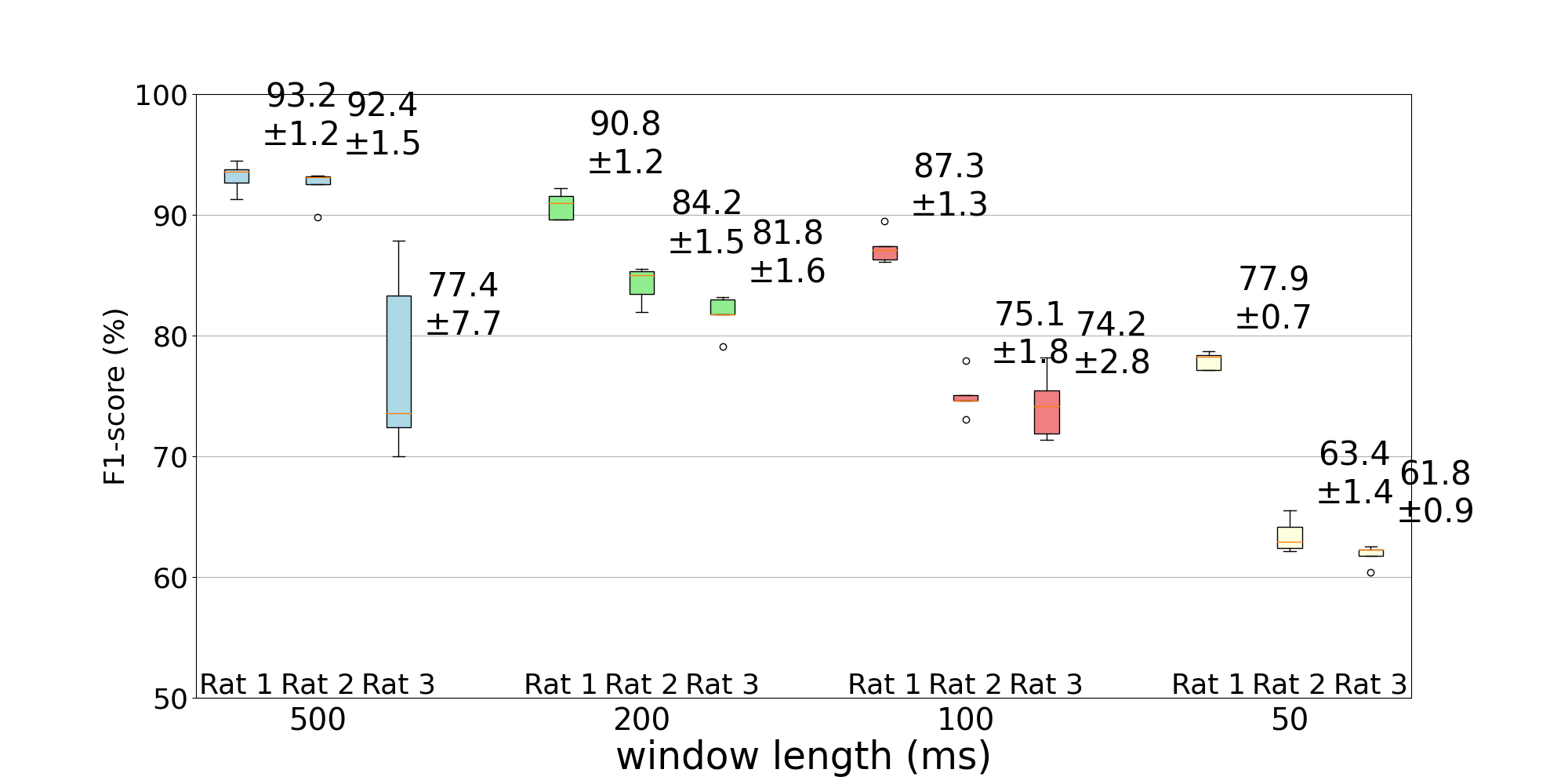}%
\label{fig_second_c}}
\caption{Accuracy (a) and F1-score (b) obtained by applying LSTM and early stopping to the studied signals.}
\label{fig12}
\end{figure*}

The LSTM network is a recurrent-based model capable of storing information for a longer period than standard recursive neural network (RNN) \cite{[24]hua2019deep}.
Several studies have successfully applied LSTM for Time Series Classification (TSC), such as separation of finger movements from EMG signal \cite{[27]millar2022lstm} or hand gesture recognition \cite{[28]jabbari2020emg}, by feeding the classifier with both raw signals and previously extracted. More recently, LSTM has been applied to classify ENG signals in~\cite{[6]Elisa}. 
In our work, we perform an investigation using this classifier. In comparison to \cite{[6]Elisa}, here the training process will be optimized by introducing, as in the other classifiers, an early stopping metric that will be detailed in the next section and defines the best period among the previous ones obtained. Also, the pipeline for generating results will be much simpler and faster.

To summarize, in the classic CNN, kernels move and extract information along two directions.  In the Inception Time CNN, we employ sub-kernels of variable, smaller lengths to perform a low-pass filtering on the signal. Subsequently, by recombining the output. The ENGNet augments the dimensionality of the system, followed by a linear combination, in order to subsequently resolve the system by identifying spikes.
The LSTM seeks temporal relationships through the utilization of intermediate frames. It assesses whether certain patterns repeat over time. The classification performance of all these networks are investigated in next section.

\section{Results and Discussions}
\label{sec:results}
In this section we present simulation results obtained using the data-set provided in \cite{ [12]Identification} and preprocessed as detailed in Section~\ref{sec:preprocessing}. 

The ENG signal was divided into temporal windows to evaluate the variability of the accuracy metric and F1-score as a function of the duration of the input ENG signal. Four different window sizes, i.e. $500$, $200$, $100$, and $50\,$ms, were chosen to analyze the prediction capabilities of the networks presented in Sec.~\ref{sec:classification}. All the networks were evaluated through a $5$-fold cross-validation set-up, in which $80$\% of the data is for training, while the remaining $20$\% for the testing. 

Compared to the original values presented in \cite{ [12]Identification}, a much lower window range is proposed due to the characteristics of the ENG signal. Since the ENG signal has a high frequency content and its smallest characteristic are the spikes that compose it, the kernel is chosen so that its value falls within the absolute refractory period of a nerve spike, lower than $1\,$ms. In this way the classifier is able to refine its network towards this type of signal allowing a classification with good levels of performance. 

A study was carried out to optimize the shutdown time of the networks using an early stopping method to eliminate problems associated with the overfitting. The early stopping method consists in the shutdown of the training when the accuracy of the validation dataset does not increase more compared with the $8$ previous epochs. About the overfitting, the networks parameters were set thanks to an ablation study to optimize the network results.

All the results are reported using F1-score. This parameter is optimal compared to the accuracy when dealing with unbalanced classes. Additionally, it allows to give more weight to false positives and false negatives obtained during the classification. Within a medical investigation, ensuring a good classification is crucial by eliminating errors caused by these secondary elements. In particular, in Figs.~\ref{fig9}-\ref{fig12} the boxplots of the various results for individual and classifier are represented.

In all the results, a reduction of the F1-score value can be observed as the duration of the observation window decreases. This is due to the reduction of the input data. Animal 1 shows best results compared to the others. This phenomenon can be due either to the bad positioning of the apparatus in the rat or simply to uncontrollable events during the acquisition phase. The signal has a strong noise component that affect the efficacy of the classification.
\begin{figure}[!t]
\centering
\includegraphics[width=\columnwidth]{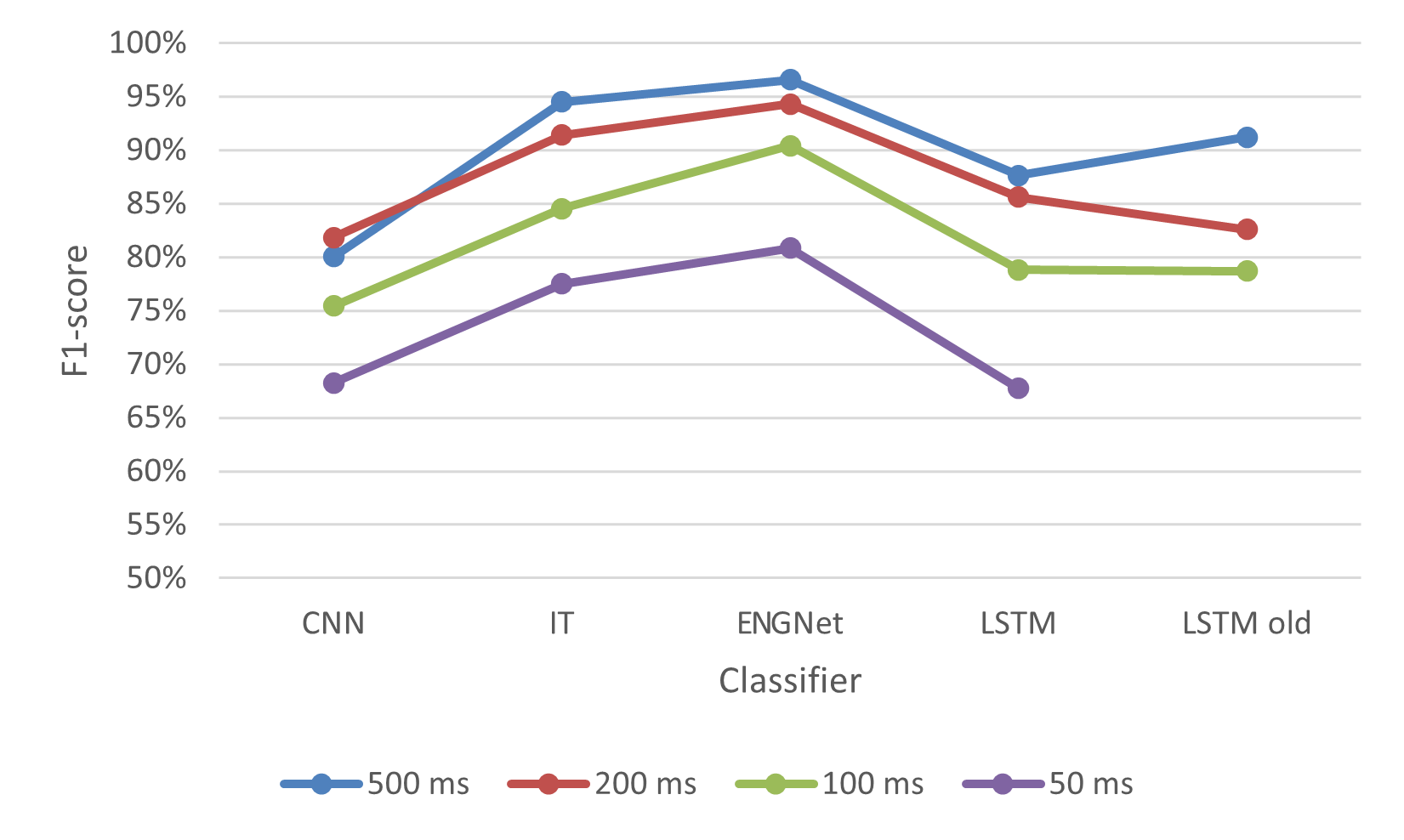}
\caption{Average F1-score values obtained for the window of the studied classifiers. "LSTM old" with ref \cite{[6]Elisa}.}
\label{fig13}
\end{figure}

The CNN network is the simplest among the studied ones and, therefore, it is able to extract less information from the recorder signals in the data sets. This consideration finds confirmation from the results reported in Fig.~\ref{fig9}. Very interesting results can be observed for IT and ENGNet These are reported in Figures~\ref{fig10} and \ref{fig11}. The accuracy is above $90$\% up to the $200\,$ms window duration for IT, and up to up $100\,$ms window duration for ENGNet. Such good results for small window sizes allow for real-time application, as the information is preserved in an optimal way despite the reduction of the data size at the input of the network. In particular, ENGNet stands out as the most suitable candidate for a real-time deployment. Regarding LSTM in Fig. \ref{fig12}, a reduced preprocessing compared to the case of \cite{[6]Elisa} leads to a reduction in the signal classification performance. For large windows, such as $500\,$ms, if the input signal is not adequately preprocessed, there is the risk of loss of information content. Indeed, the memory contribution that is exploited by the network can be lost due to noise. As a consequence, a reduction of the F1-score value is observed. For the other two windows compared the value remains almost constant because the input windows are smaller and more comparable with the size of the kernels used to evaluate the temporal correlations set to $20\,$ms. An improvement in performance could be obtained by studying this feature but increasing the total number of parameters. The LSTM does not show performance improvement in $50\,$ms window because of the size similarity with the kernel. For this reason and for its higher computational complexity, the LSTM approach can be considered unsuitable for real-time classification when small input windows are used.

\begin{figure}[!t]
\centering
\includegraphics[width=\columnwidth]{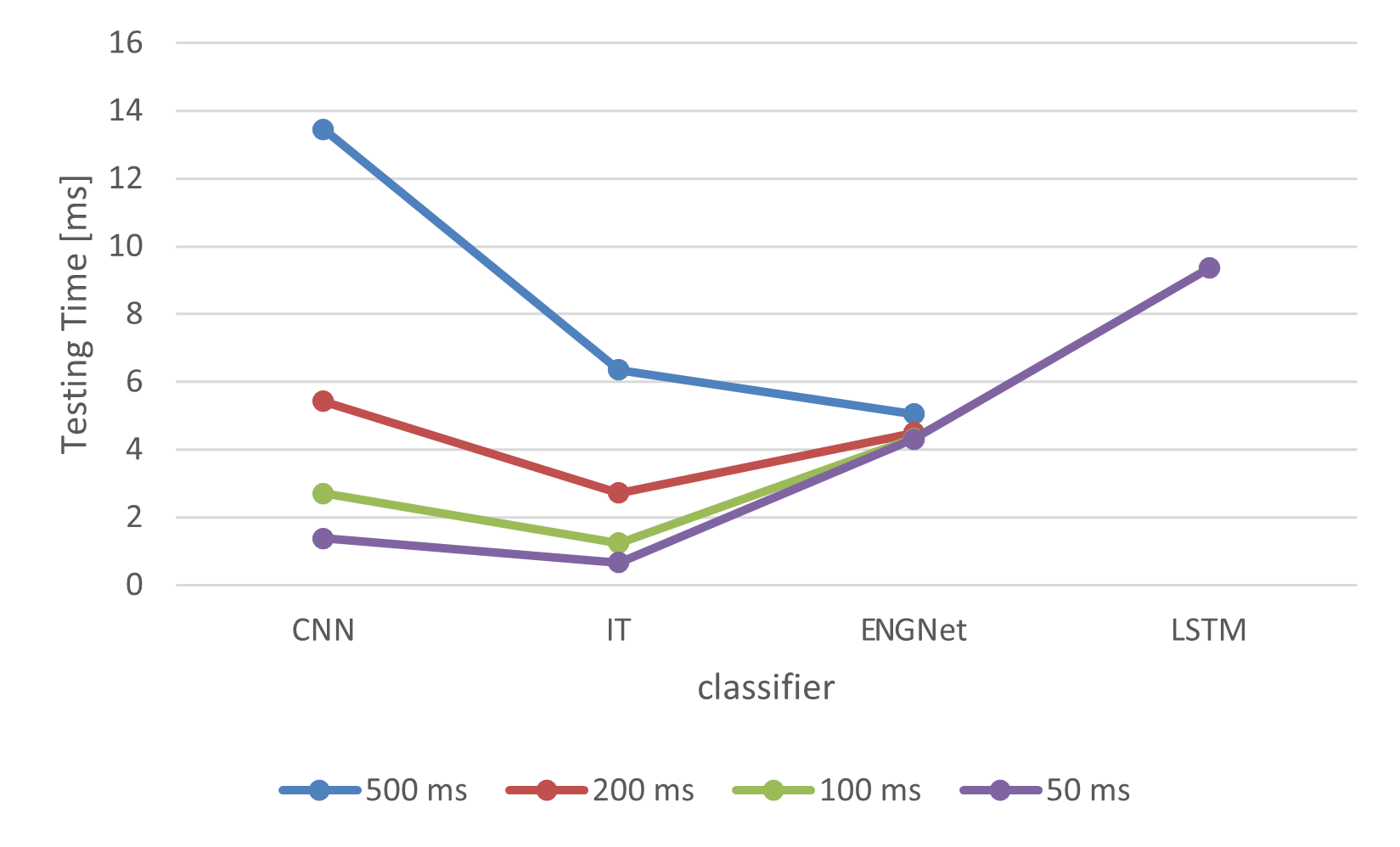}
\caption{Average test time values per sample per window of the studied classifiers useful for real time applications.}
\label{fig14}
\end{figure}

Figure~\ref{fig13} reports the average values of F1-score as a function of the input window used. ENGNet is distinguished by its classification performance.  In particular, the $100\,$ms and $200\,$ms windows could be appropriate for real-time applications. 
The test time can also be used as performance metric to evaluate whether or not a classifier can be used in real-time ND\&S systems. In Fig.~\ref{fig14}, it is shown how the test time impacts the classification. 
The CNN shows good timing characteristics for low window values reaching 1.38ms per classified item. The IT is the best performing classifier at the lowest windows hitting 0.76 ms per classified item. 
The behavior of the ENGNet is different. In fact, the time parameter is mainly constant for different window size and between 4-5ms for classified element. Compared to the other classifiers, ENGNet time performance not depend with the window size. 
As far as LSTM is concerned, the figure reports the value associated with 50ms of window as it is the only one relevant for real time applications. 
The other values are excessively high, confirming that LSTM can be discarded as solution.
\begin{figure}[!t]
\centering
\includegraphics[width=\columnwidth]{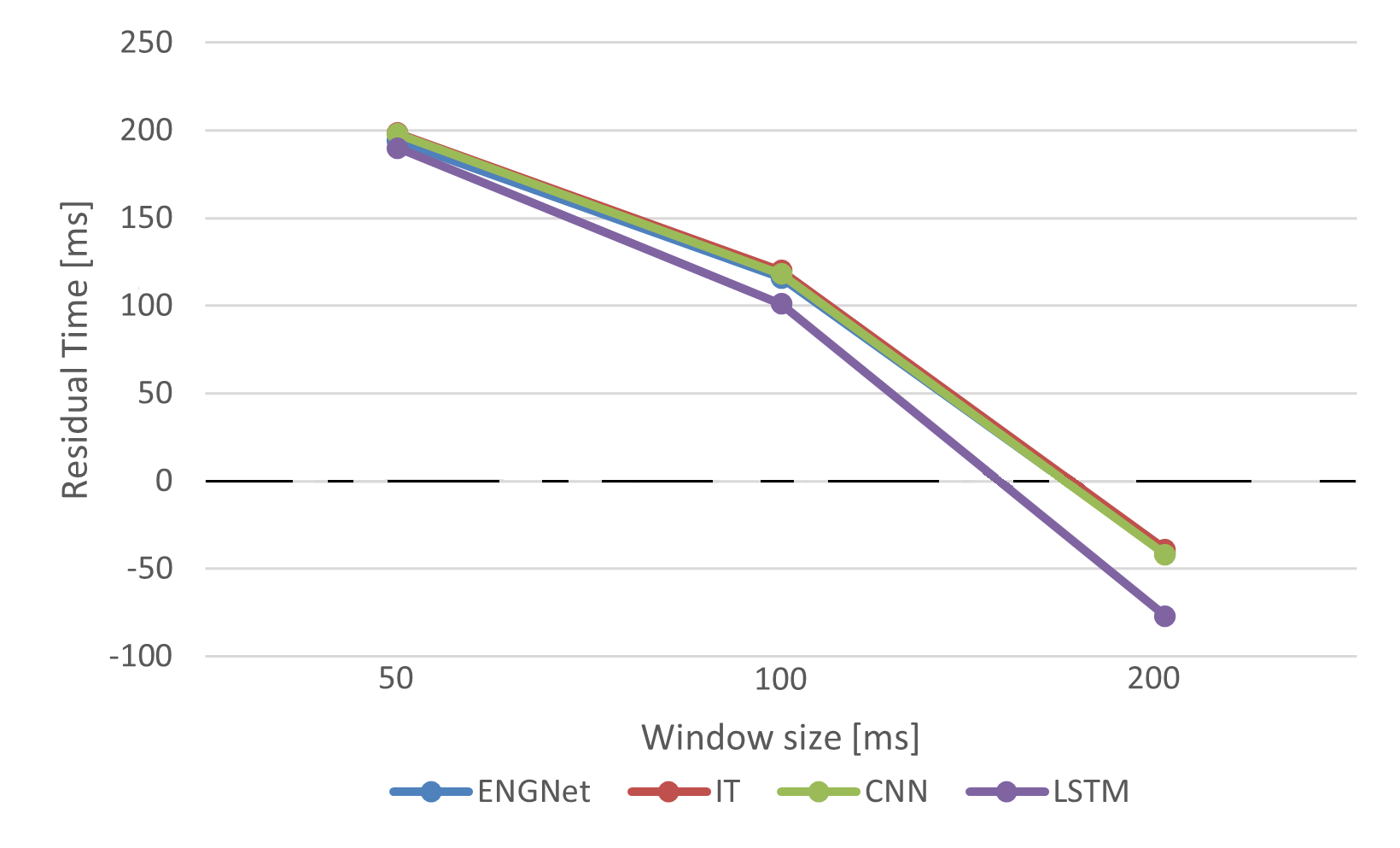}
\caption{Residual time for implementing classification as a function of the window size. The black dashed lines gives the acceptance threshold.}
\label{residualtime}
\end{figure}
\begin{figure}[!t]
\centering
\includegraphics[width=\columnwidth]{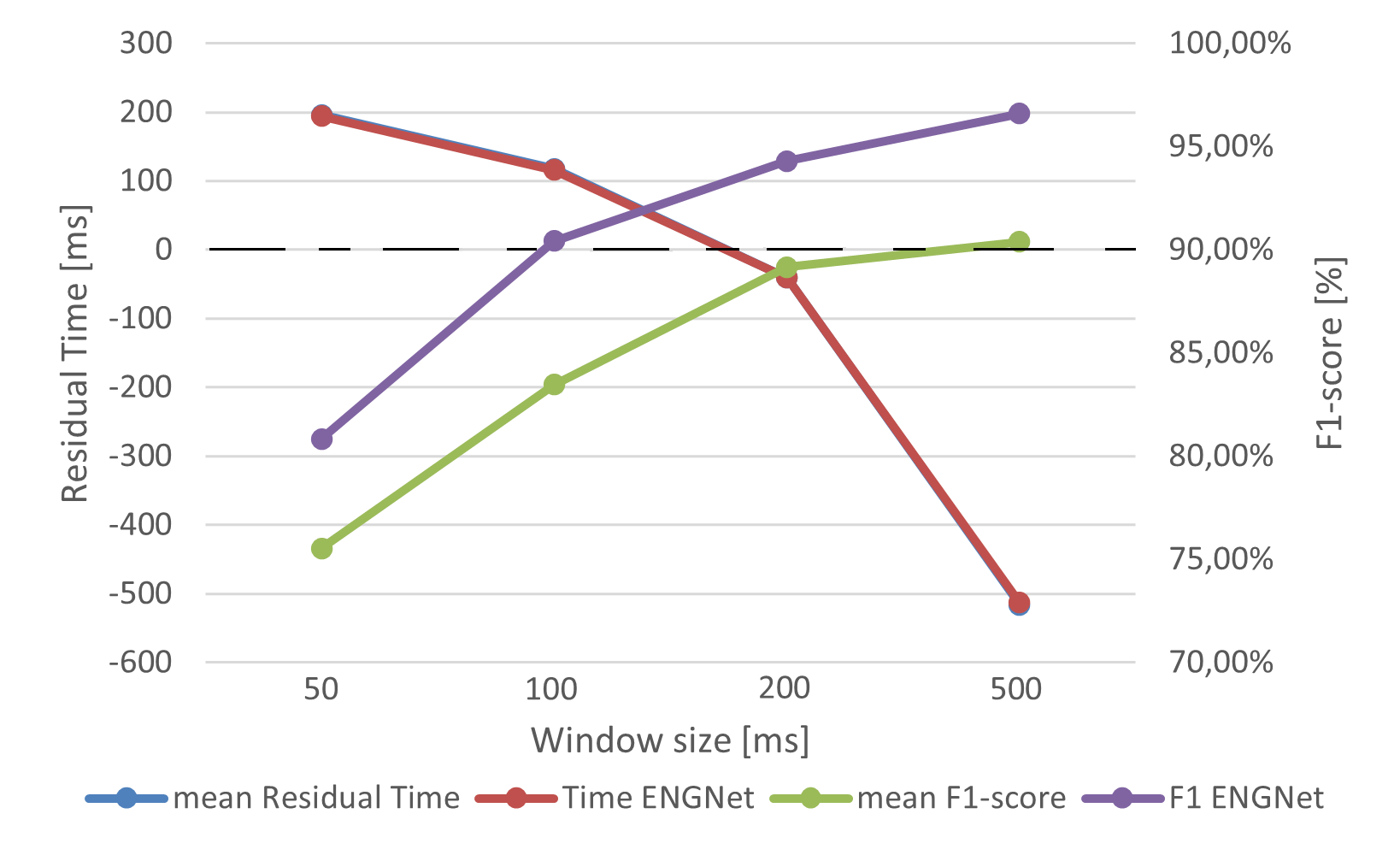}
\caption{Trade-off between the parameter residual time and F1-score, in black the acceptance threshold.}
\label{tradeoff}
\end{figure}

Figure \ref{residualtime} reports the residual time for classification as a function of the windows size. The windows of duration $500\,$ms has been not reported since it exceeds the $300\,$ms human percpetion delay. From the figure it can be easily seen for which windows real-time operations is possible. CNN, IT, and ENGNet all fall within the defined threshold for windows less than $100\,$ms, showcasing their suitability for they real time  implementation. Once again LSTM presents the worst results. From this we can say that the proposed implanted device can be used to perform ENG measurements.

From the results reported in Figs.~\ref{fig13}, \ref{fig14}, and \ref{residualtime}, we can note a presence of trade-off between the residual time and the F1-score as a function of the window size. This trade-off is better illustrated in Fig.~\ref{tradeoff}. Looking at the average value\footnote{The LSTM results were excluded during the mean calculation due to their obtained poor performance.}, a contrasting behavior of the physiological time requirement and of the classification score can be observed. Focusing the attention on the ENGNet, i.e. the one with the best classification performance, we can note that for a window of $100\,$ms the residual time can stay above the threshold, while maintaining in the same time a F1-score value upper then $90\,\%$. For this reason we can say that, between the network studied, the ENGNet using a window of $100\,$ms is the most suitable one in terms of real-time ENG data classification.

An essential consideration to assess the network feasibility is the computational load and potential device implementation, which can be estimated through the total number of network parameters. Specifically, the parameter counts for each network are as follows: 1.242.116 for LSTM, 577.956 for CNN, 50.948 for IT, and 4.964 for ENGNet. This underlines that LSTM's excessive parameters hinder real-time implementation, while ENGNet stands out as the optimal choice with the significantly lowest count of parameters.

An ablation study was conducted to analyze input features and individual components defining classifiers. Three approaches were proposed for input characteristics based on the multi-contact geometry of a cuff electrode. Longitudinal, transversal, and random electrode ordination have been done. The study revealed the use of 1D kernels ensures performance invariance with sorting type and reduces computational time, as report in Fig. \ref{fig14}. About the classifiers component, we report the final optimal configuration for each model. For ENGNet, an investigation on optimal network parameter size emphasized the importance of the first line of kernels and identified $100\,$ms windows as the best performing for real time application. Other parameters minimally affected the accuracy. IT utilized $3$ kernels per block to emphasize specific spectral characteristics, with different kernel sizes studied. For IT, ENGNet, and CNN the convolution process involved iterative block additions until no performance improvement was observed over $5$ cross-validations, determining the optimal number of repetitions. LSTM study was not performed due to its identified non-optimality for this signal type. Increasing its parameters for better performance would hinder implementation on microcontrollers with limited memory and increase execution time so high, as reported in Fig. \ref{fig14}.

\section{Comparison with the existing literature}

The networks used in our study were originally designed for different signals, such as EMG (electromyographic, muscle) or EEG (electroencephalographic, brain) \cite{[16]ismail2020inceptiontime, [23]feng2022efficient, [24]lstm}. These networks have been profoundly modified to be adapted to the analysis of ENG signals. Due to the intrinsic diversity between these signals, in terms of characteristics and information content, a comparison between the previous models and the new ones would be inconsistent and meaningless. Since the results are not comparable they were not reported in the paper. 

However, a comparison is made with other networks applied to ENG signals with the state of the art. The results of our method, reported in Fig. \ref{fig13} are comparable or superior to those reported in the literature, with improved performance compared to \cite{[6]Elisa, [7]Federica,[14]koh2020selective}. 
Even using smaller windows, we achieved superior results. The spiking approach of \cite{[7]Federica} is interesting, and if improved could lead to significant results. is able to obtain good results on 10 classes, more numerous than ours, which concerns different stimuli with different intensities. Our classifier shows ability in general signal classification, but faces difficulty in defining the intensity. A hybrid approach could improve performance in both aspects. The spiking approach of \cite{[14]koh2020selective}, although it has not reached the same performance levels as \cite{[7]Federica}, presents an interesting and easily implementable signal encoding for a hybrid approach. Regarding \cite{[6]Elisa}, the use of LSTM, similar to ours, did not obtain comparable results with ENGNet and IT, confirming that there are more efficient networks than LSTM with lower computational weight. Finally, \cite{newcastle2020} obtained excellent results, sometimes superior to ours. The use of machine learning techniques could be interesting to improve the classification, but it should be considered that the classification of \cite{newcastle2020} concerns similar intensities of stimuli. Therefore performance could vary with different stimuli, as in our case. The excellent results of \cite{newcastle2020} and the low computational complexity deserve further studies to verify whether they can be maintained with datasets similar to those of \cite{[7]Federica} or ours.

\section{Conclusion}
\label{sec:conclusion}

This paper focuses on the real-time ANN-based classification of sensory/motor stimuli carried by ENG signals. Four different ANN-based approaches are proposed and evaluated on ENG signals measured by means of PNIs. All of these take into account the human sensory response limit of $300\,$ms. The proposed approaches are useful for the implementation of neural decode and forward (ND\&S) systems, where the intended stimulus contained in the ENG signal must be relayed after nerve injury to support the recovery of a sensory/motor function. A possible ND\&S architecture is proposed, where ANN-based classification is performed by an external PU. The design of ANN-based classifiers relies on a MIMO ENG signal modelling, which is another contribution of this work. 
The evaluation over an available data set has revealed interesting insights with different time windows. Among the proposed approaches, the ENGNet architecture, derived from EEGNet for classification of EEG signals, emerges as a preferred choice for real-time applications, balancing notable performance with temporal efficiency. On the other hand, LSTM, due to its computational complexity, reveals significant limitations, underscoring the need for a thoughtful consideration of complexity in real-time classification applications.

Other classifiers of different types could be studied and compared with the collection of convolutional networks proposed in this work. In fact, as a result of future discoveries on the physiological mechanisms of how the nervous system works, new classification systems could be generated, such as spiking neural networks that will attempt to mimic nervous behavior to correctly encode the information contained.

\ifCLASSOPTIONcaptionsoff
  \newpage
\fi
\footnotesize
\bibliographystyle{IEEEtran}
\bibliography{biblio.bib}

\begin{IEEEbiography}[{\includegraphics[width=1in,height=1.25in,clip,keepaspectratio]{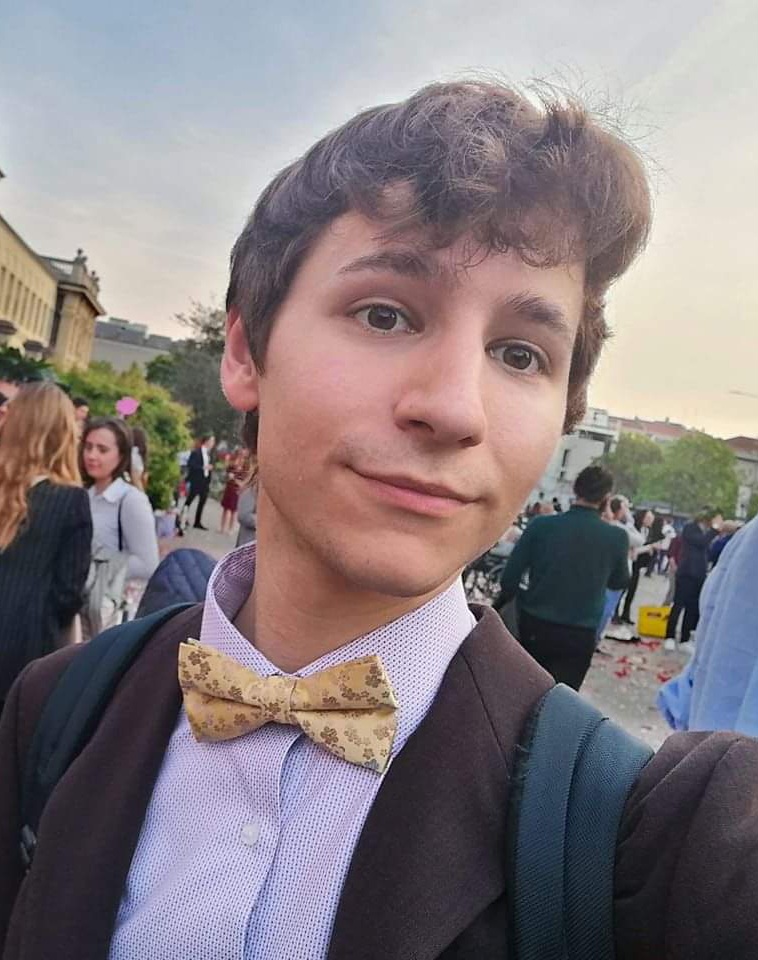}}]{Antonio Coviello} (Student Member, IEEE) is currently a PhD student in Telecommunication Engineering at Politecnico di Milano, Milan, Italy at the Dipartimento di Elettronica, Informazione e Bioingegneria, Politecnico di Milano (DEIB). He is graduated in Biomedical Engineering at Politecnico di Milano (MSc in 2021). His main research interests include the design of implantable medical devices, bioelectronics, data analyst, communication systems, 3D printing, finite element COMSOL simulation, intrabody comunication device, wearable sensors. 
\end{IEEEbiography}

\begin{IEEEbiography}[{\includegraphics[width=1in,height=1.25in,clip,keepaspectratio]{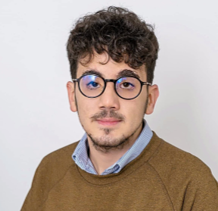}}]{Francesco Linsalata} (Member, IEEE)  received PhD and M.Sc. degrees cum laude in Telecommunication engineering from Politecnico di Milano, Milan, Italy, in 2019 and 2022, respectively. He is a researcher at the Dipartimento di Elettronica, Informazione e Bioingegneria, Politecnico di Milano.  His main research interests focus on V2X communications and waveforms design for B5G wireless networks. He was the co-recipient of the best-paper award and recipient of the best student paper award at BalkanCom'19. 
\end{IEEEbiography}

\begin{IEEEbiography}[{\includegraphics[width=1in,height=1.25in,clip,keepaspectratio]{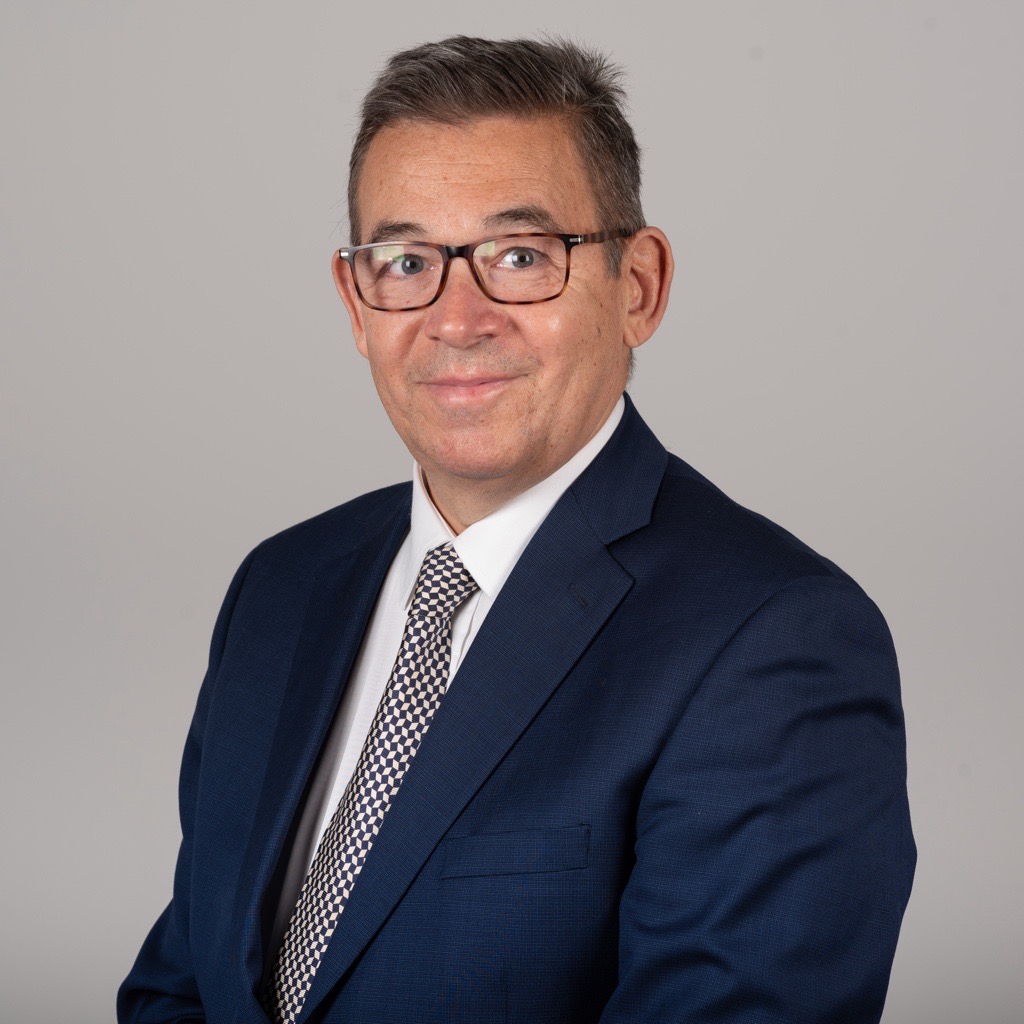}}] {Umberto Spagnolini} (Senior Member, IEEE) is Professor of Statistical Signal Processing, Director of Joint Lab Huawei-Politecnico di Milano and Huawei Industry Chair. His research in statistical signal processing covers remote sensing and communication systems with more than 300 papers on peer-reviewed journals/conferences and patents. He is author of the book Statistical Signal Processing in Engineering (J. Wiley, 2017). The specific areas of interest include mmW channel estimation and space-time processing for single/multi-user wireless communication systems, cooperative and distributed inference methods including V2X systems, mmWave communication systems, parameter estimation/tracking, focusing and wavefield interpolation for remote sensing (UWB radar and oil exploration). He was recipient/co-recipient of Best Paper Awards on geophysical signal processing methods (from EAGE), array processing (ICASSP 2006) and distributed synchronization for wireless sensor networks (SPAWC 2007, WRECOM 2007). He is technical experts of standard-essential patents and IP. He served as part of IEEE Editorial boards as well as member in technical program committees of several conferences for all the areas of interests.
\end{IEEEbiography}

\begin{IEEEbiography}[{\includegraphics[width=1in,height=1.25in,clip,keepaspectratio]{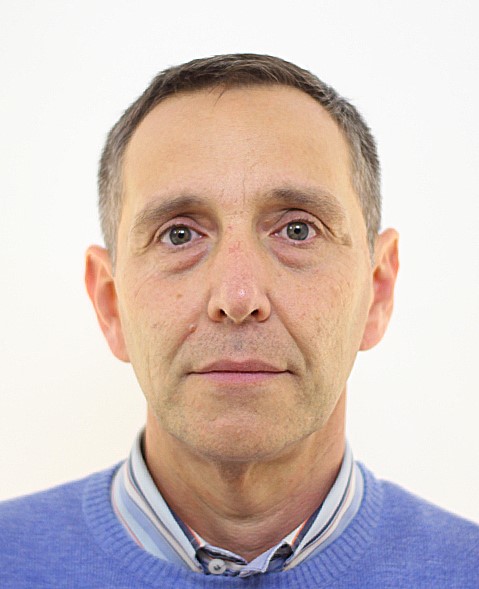}}] {Maurizio Magarini} (Member, IEEE)  received the M.Sc. and Ph.D. degrees in electronic engineering from the Politecnico di Milano, Milan, Italy, in 1994 and 1999, respectively. In 1994, he was granted the TELECOM Italia scholarship award for his M.Sc. Thesis. He worked as a Research Associate in the Dipartimento di Elettronica, Informazione e Bioingegneria at the Politecnico di Milano from 1999 to 2001. From 2001 to 2018, he was an Assistant Professor in Politecnico di Milano where, since June 2018, he has been an Associate Professor. From August 2008 to January 2009 he spent a sabbatical leave at Bell Labs, Alcatel-Lucent, Holmdel, NJ. His research interests are in the broad area of communication and information theory. Topics include synchronization, channel estimation, equalization and coding applied to wireless and optical communication systems. His most recent research activities have focused on molecular communications, massive MIMO, study of waveforms for 5G cellular systems, vehicular communications, wireless sensor networks for mission critical applications, and wireless networks using unmanned aerial  vehicles and high-altitude platforms. He has authored and coauthored more than 120 journal and conference papers. He was the co-recipient of two best-paper awards. He is an Associate Editor of IEEE Access, IET Electronics Letters, and Nano Communication Networks (Elsevier). He has been involved in several European and National research projects.
\end{IEEEbiography}

\end{document}